%% file: main.tex
\documentclass[a4paper,fleqn]{cas-dc}

\usepackage[numbers]{natbib}
\usepackage{lineno}
\usepackage{xcolor}
\modulolinenumbers[5]
\usepackage{algpseudocode}
\usepackage{hyperref}
\usepackage{algorithm}
\usepackage{algpascal}
\usepackage{mathtools}
\usepackage{caption}
\usepackage{nomencl}
\usepackage{ntheorem}   
\usepackage{lipsum}     
\usepackage{graphicx}
\usepackage{bm}

\DeclareRobustCommand{\uvec}[1]{{%
		\ifcsname uvec#1\endcsname
		\csname uvec#1\endcsname
		\else
		\bm{\mathbf{#1}}%
		\fi
}}
\def\tsc#1{\csdef{#1}{\textsc{\lowercase{#1}}\xspace}}
\tsc{WGM}
\tsc{QE}
\tsc{EP}
\tsc{PMS}
\tsc{BEC}
\tsc{DE}

\graphicspath{{graphics/}, {manuscripts/}}

\newcommand{\makehighlight}[1]{\textcolor{black}{#1}}

\begin{document}

\title [mode = title]{Line-Circle-Square (LCS): A Multilayered Geometric Filter for Edge-Based Detection}                      



\author[1]{Seyed Amir Tafrishi}[type=editor,
                        auid=000,bioid=1,
                        orcid=0000-0001-9829-3144]
\cormark[1]
\ead{amir@ce.mech.kyushu-u.ac.jp}


\address[1]{Department of Mechanical Engineering, Kyushu University, Kyushu, Japan}

\author[2]{Xiaotian Dai}[orcid=0000-0002-6669-5234]
\ead{xiaotian.dai@york.ac.uk}
\address[2]{Department of Computer Science,
 University of York, York, UK.}

\author[3]{Vahid Esmaeilzadeh~Kandjani} 

\ead{v.esmaeilzadeh@gmail.com}

\address[3]{Department of Computer Engineering, University of Tabriz, Aras Free-Zone Campus, Iran}


\cortext[cor1]{Corresponding author}


\begin{abstract}
This paper presents a state-of-the-art filter that reduces the complexity in object detection, tracking and mapping applications. Existing edge detection and tracking methods are proposed to create suitable autonomy for mobile robots, however, many of them face overconfidence and large computations at the entrance to scenarios with an immense number of landmarks. The method in this work, the Line-Circle-Square (LCS) filter, claims that mobile robots without a large database for object recognition and highly advanced prediction methods can deal with incoming objects that the camera captures in real-time. The proposed filter applies detection, tracking and learning to each defined expert to extract higher level information for judging scenes without over-calculation. The interactive learning feed between each expert increases the consistency of detected landmarks that works against overwhelming detected features in crowded scenes. Our experts are dependent on trust factors' covariance under the geometric definitions to ignore, emerge and compare detected landmarks. The experiment validates the effectiveness of the proposed filter in terms of detection precision and resource usage in both experimental and real-world scenarios.
\end{abstract}




\maketitle

\input{contents/intro.tex}
\input{contents/review.tex}

\input{contents/lcs-overview.tex}
\input{contents/lcs.tex}

\input{contents/evaluation.tex}
\input{contents/conclusion.tex}


\printcredits

\bibliographystyle{cas-model2-names}

\bibliography{referencesDIF}

\end{document}

%% file: contents/intro.tex
 
\section{Introduction} \label{sec:introduction}

Object detection and tracking are among the challenging tasks in the field of robotics and computer vision \cite{wu2015object}.
One of the research challenges is to have satisfactory real-time performance and run-time efficiency. 
In recent years, a considerable large number of detection and tracking algorithms have been proposed \cite{geronimo2009survey,luo2014multiple,zhao2019object}, some of which can reduce computation complexity and/or increase the rate of object detection \cite{TLD2012,Visiononl2014}. These detectors use real-time data to process during their activation states.
For autonomous mobile robots including autonomous driving cars, it is vital to assure fast responses during their motion. 
Additionally, due to size and energy constraint of embedded computing platforms, it is an urge to reduce resource usage and the number of sensors during active times \cite{Visiononl2014}.

To address these issues, a multi-layer filter is constructed in this paper, namely the \textit{Line-Circle-Square (LCS) filter}, for object detection using information from a two-dimensional image sensor, e.g., a camera.
The LCS filter is a multilayered geometric filter for edge-based detection. \makehighlight{ In our previous work, we did a preliminary study on the \textit{Line-Circle filter} \cite{tafrishi2017line}, the predecessor of the LCS filter. The algorithm of the LC filter was run offline with limited performance evaluation in the scene. Compared with the LC filter \cite{tafrishi2017line}, we have optimized computation flow, restructured the algorithmic conditions and improved the code for real-time applications in all of the experts. The LCS filter is advanced with more interactive information transfer between each expert. Also, the proposed filter has an extra layer --- the \textit{Square expert}, in contrast to the Line-Circle (LC) filter.} By utilizing inertial measurement units (IMUs) in parallel to the real-time captured images, an estimated scene is built that predicts possible objects that are passing or moving towards the robot. The proposed filter optimizes the data collection from existing edge detection methods \cite{rosten2005fusing,Rosten2006} or event-based camera \cite{EVENTBASECAMERA} to ignore/concentrate on particular landmarks by using geometric and kinematics conditions.
Besides these capabilities, with inspiration from certain estimators, e.g., the \textit{Extended Kalman Filter (EKF)} \cite{davison2007monoslam,TommyIROS2015}, errors are minimized to prevent any misclassification or integrated uncertainties of detected landmarks.

\makehighlight{Compared with other related filters in the literature, the proposed LCS filter has the following unique advantages: (1) it reduces computation demand; (2) it has the ability to minimize the problem of overconfidence during detection; (3) real-time process for detecting abnormal behaviors at outside world such as partial detection of incoming objects toward the camera/moving vehicle; (4) primary detection with geometrical computation which creates a different level of information, i.e., low (edges) to high (layers) for mapping and localization; (5) the multi-layer nature makes it suitable for real-time processing with potential to be executed in parallel.
In contrast to other learning methods, e.g., deep learning \cite{wang2017deepvo,zhan2018unsupervised}, the LCS filter is an online learning algorithm, without the need for a large pool of training images and data.
For implementation, this geometric filter only requires IMU and camera for working; Hence, it can work in battery-powered and memory-constrained systems including autonomous vehicles. Also, we have made the source code of LCS filter public for the benefit of the industrial and research communities}\footnote{GitHub Source: \url{https://github.com/SeyedAmirTafrishi/LCS_Filter/tree/LCS_RAS}. }.

\makehighlight{The proposed LCS filter works by collecting the detected edges (features) from the monocular camera. Next, Line expert groups and removes the detected features depending on the updates that come from Circle and Square experts. This operation decreases the computation size while preparing the features for higher accuracy of matching depending on the number of captured landmarks in different parts of the frame. Then, the Circle expert estimate features and do kinematic grouping for finding layers (grouped features with the same properties). Also, this expert predicts the features and layers through the motion field kinematics by using the angular rotation and velocity of the vehicle's IMU sensor. Square expert finally does the geometric and kinematic matches of the circles to construct the highest level of information as the layer of the objects in the scene. These kinematic classifications in the Circle and Square help us to find independently moving objects in the scene and have more consistent and accurate feature detection.} 

The paper is organized as follows: the related work is first given in Section~\ref{sec:review}. An overview of the LCS filter is introduced in Section~\ref{sec:overview}. Then, the LCS filter is decomposed into three different subsections: line, circle, and square in Section~\ref{sec:lcs-filter}. Evaluations in both simple and real-world scenarios are given in Section~\ref{sec:evaluation}, followed by discussions and conclusions in Section~\ref{sec:conclusion}.

%% file: contents/review.tex
\section{Related Work} \label{sec:review}

Object detection and tracking methods have been widely studied \cite{geronimo2009survey,luo2014multiple,zhao2019object} to create accurate and fast analysis over the surrounding environment. In this section, we give a brief discussion for existing methods and highlight the difference in contrast to the LCS filter.


 \subsection{Object Detection}
The primary goal of object detection is to localize the existing objects in the captured scene from input images. Note that objects can correspond to many combined class of objects or to a single one. To determine the features of surrounding, edge detection methods \cite{Rosten2006,tuytelaars2008local} have been the most common studies. Also, there have been many alternative methods. As a classic approach for understanding the environment, the stereo-based model was firstly designed by Murray and Little for SLAM \cite{Murray2000}. 
Other detection methods were studied related to the point analysis \cite{Sand2008} and image contouring \cite{bibbyeccv08}. 
Next, a semi-dense monocular SLAM with the integration of color was applied to determine the surrounding objects at MIT \cite{PillaiRSS2015}. An overall review was presented by Sun et al. about the vision-based localization methods for other vehicles on the road \cite{VehicleREview2006}. There is another survey that summarized different approaches based on edge detection for object/place recognition \cite{williams2009comparison}. This study showed that image-to-image matching, based on the appearance in the surrounding, has better scaling. In particular, when the environment is large, the map-to-map or image-to-map approaches face limitations. However, if a overwhelming number of landmarks is detected, the loop closer 
and effective recognition in these SLAM problems will be impractical for its applications \cite{mur2015orb,mur2017visual,williams2009comparison}.   

\subsection{Corner Detection and Applications} 
Rosten and Drummond proposed a machine learning corner detection method \cite{Rosten2006} as an efficient feature detection of the environment in real-time. Next, the moncular SLAM studies were taken place by applying different approaches for mapping based on feature detection \cite{civera2008inverse,davison2007monoslam,eade2006scalable}. 
%
Monocular SLAM with corner detection, as the most practical and advanced method, was analyzed to have better solutions for overconfidence and dealing with high computations related to the detected landmark.
Lui and Drummond proposed a new system for constant time monocular SLAM that uses only 2D measurement and takes the image graph with sparse pairwise geometries. There were some improvements such as no global consistency and bundle adjustments but the system was based on a multi-camera perspective and it was also dependent on a reference image \cite{Tomy2015ICRA}. In the research field of edge detection, a fast event-based corner detection method is also proposed in \cite{mueggler2017fast}. This detection method is based on a novel event-based camera that only responds to local changes in brightness, and can detect edges directly from the camera. The other work is related to Kalman filter vision \cite{liu2018collision}, which has low complexity but can still be overwhelmed by increasing the number of landmarks.


\subsection{Detection and Tracking Filters}
The Kalman filter reduction was applied for SLAM problems for using bundle adjustment by sparse matrix and double windowing method by Gamage and Drummond~\cite{TommyIROS2015}. They tried to decrease the dimension of the covariance matrices of the camera and landmark position. Although it minimized some non-linearities that create inconsistency in extended Kalman filter (EKF) \cite{davison2007monoslam}, it was not able to deal with overwhelming landmarks.
A requirement for the solely image-based filter as a superior method rather than a classic Kalman filter has been remained unsatisfactory. Vision dependent applications require a filter that is able to do real-time computation reduction within diverse machine learning evaluation layers. Also, KF and EKF will reach overconfidence and might take the noise as an actually existed object.

On the other hand, the Correlation Filter trains a linear template that can discriminate between images and their translations, which is formulated in the Fourier domain that provides a fast solution for object tracking \cite{henriques2014high, valmadre2017end}.
A Kernelized Correlation Filter (KCF) is derived in \cite{henriques2014high} that has the exact same complexity as its linear counterpart for kernel regression.
In \cite{valmadre2017end}, fully-convolutional Siamese networks are used which enables learning deep features that are tightly coupled to the Correlation Filter.
However, the correlation filter only works with a small amount of data in the training process, and it could not work if the environment changes stochastically, for example, if pedestrians pass in front of an autonomous car.
In the literature of convolutionary neural network for machine vision, various techniques existed in complexity reduction for visual detection \cite{cheng2017survey}. Some of the representative work include model compression \cite{chen2015compressing, zhang2020pruning}, quantization including binary neural networks (BNNs) \cite{rastegari2016xnor}, network pruning \cite{hu2016network} and approximated computation \cite{ma2020axby}. However, these works are limited to the context of CNNs, and cannot be applied to more general cases.

\vspace{0.5em}
\makehighlight{The Line-Circle-Square (LCS) introduced in this work uses geometrical information to filter overwhelmed data points which makes it computational efficient.
The edges are filtered and only the edges that are important or new need to be processed in a way that significant computation can be saved.
This has fundamental differences to other detection works as LCS focuses on reducing complexity and thus computational and memory cost. 
These make it an ideal front end of other object detection methods. The LCS filter extends the in-hand information level from low (edges) to high (layers). This gives more flexibility to address SLAM problems. A various of recognition algorithms can run faster because we are able to create kinematic labels for the chosen desired objects. Finally, the LCS filter is a multilayered approach that helps us to utilize parallel processing for each dedicated expert (line, circle and square). }

%% file: contents/lcs-overview.tex
\section{Line-Circle-Square (LCS) Filter} \label{sec:overview}

\begin{figure}
\centering
\includegraphics[width=3.3 in]{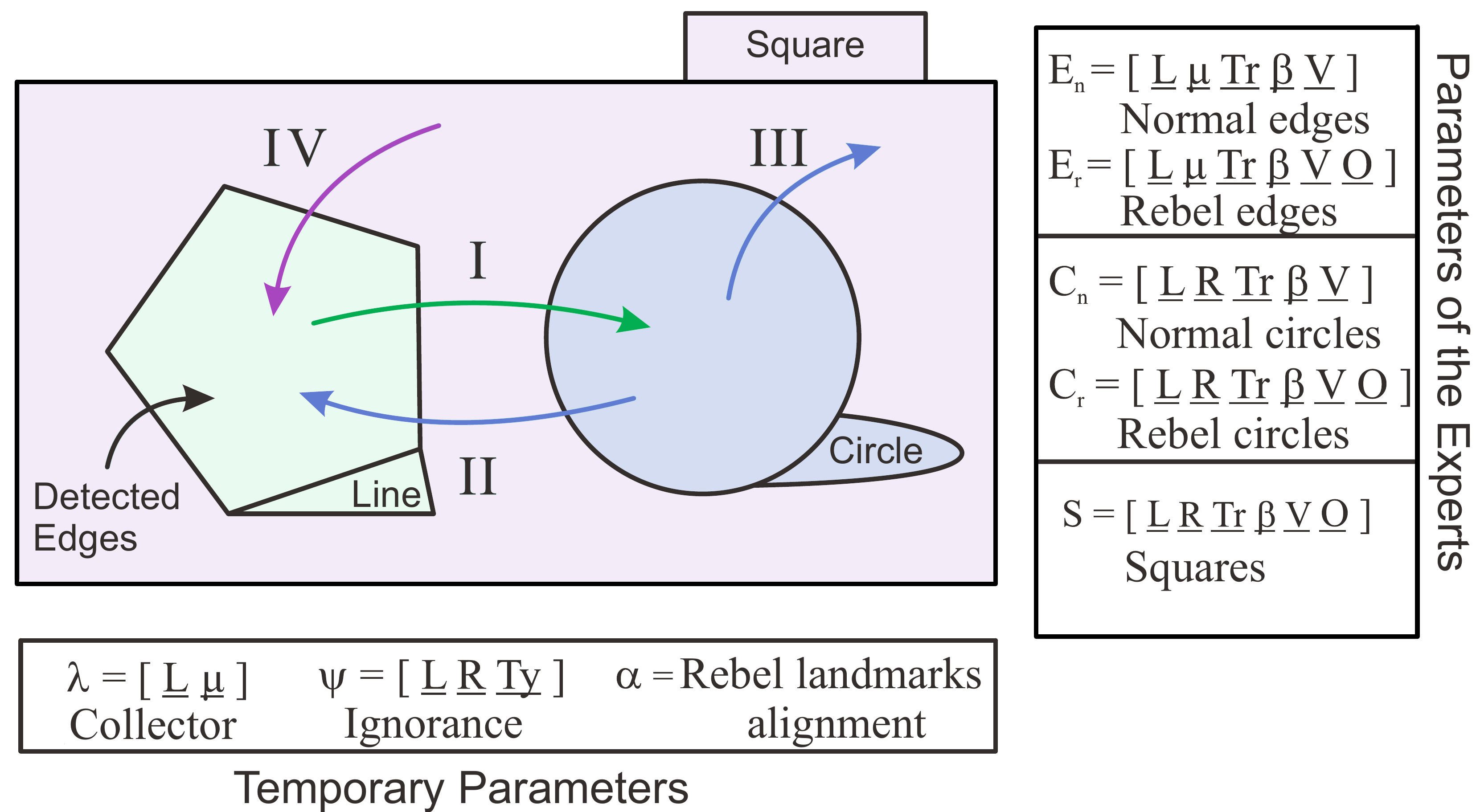}
\caption{An overview of the geometric LCS filter with data flow and filter parameters. }
\label{Fig:AlgorithmFunctioningMAP}
\end{figure}

A filter is proposed in this work of which the mainframe system has its evaluation from the surroundings in real-time. The algorithm detects objects in layers with using the past and current data from camera and IMU sensor which is obtained collectively. This filter runs without any pool of images or information except the initialization in the first frame. The novelty of this approach stands for flexible multi-level analysis on captured edge from the corner detection method \cite{rosten2005fusing,Rosten2006}. In other words, the LCS filter carries its feature analysis with filtering the data in each state of Line, Circle, Square experts. These experts update the data for future incoming frames besides correcting counterpart experts; Hence, each expert provides high-level geometrical detection from the environment. Thus, the collected edges from the camera are transferred between experts for detection, learning and tracking in each stage as Figure \ref{Fig:AlgorithmFunctioningMAP}.

\subsection{Overview of the Filter}

\begin{figure}
\centering
a)\quad\quad\quad \includegraphics[width=1.8 in, height= 2.0 in]{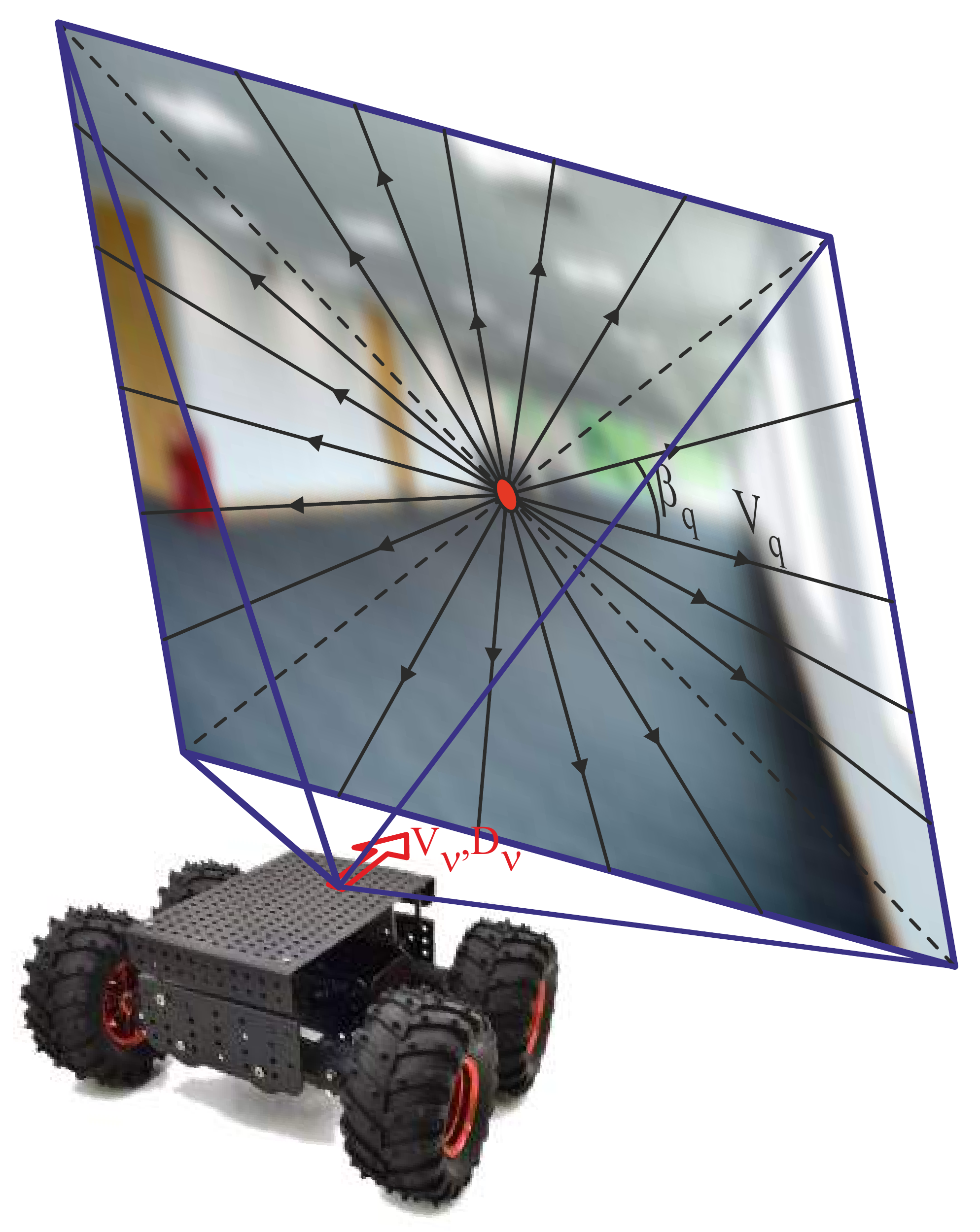}
b)\includegraphics[width=2.4 in]{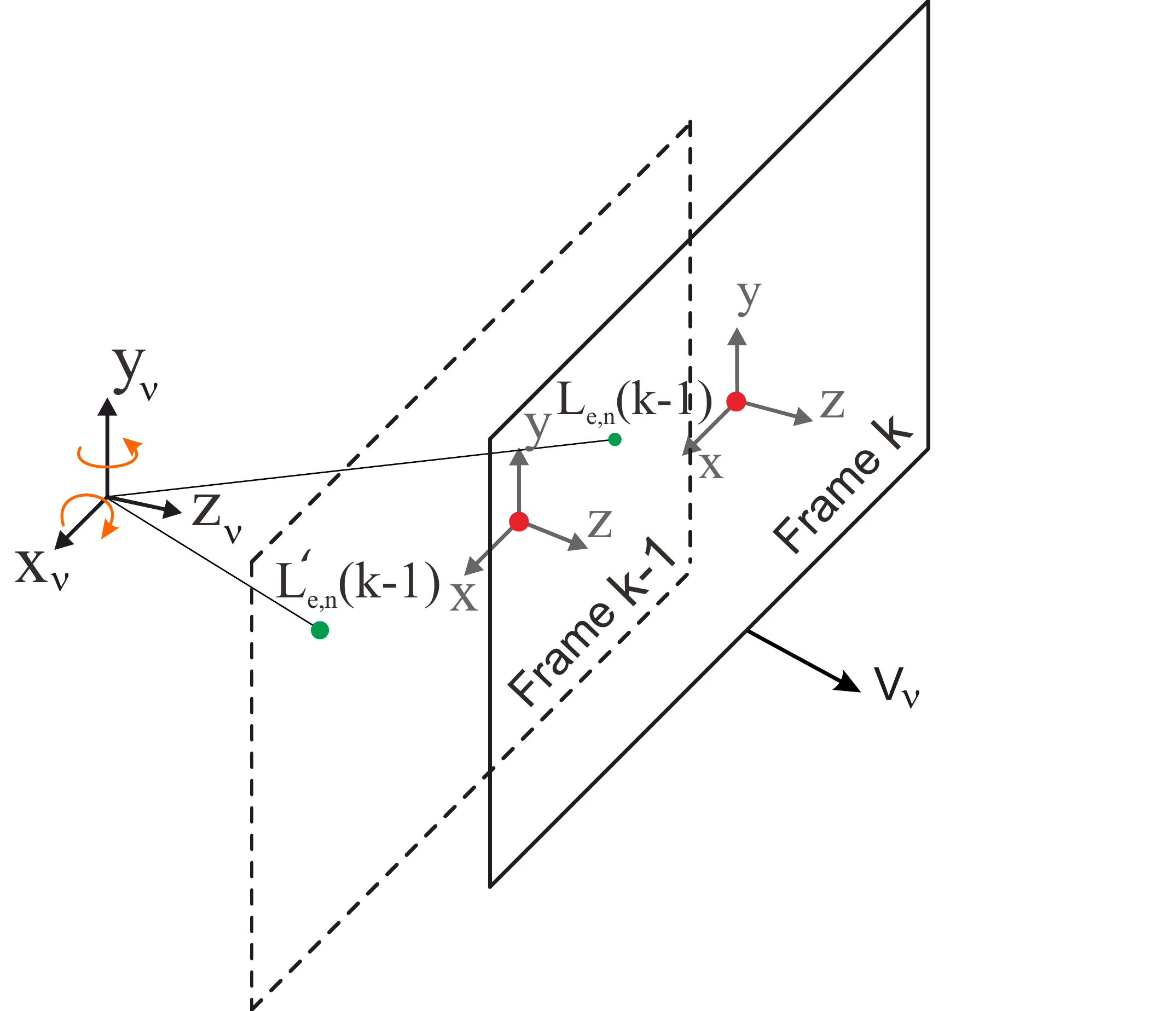}
\caption{Kinematic analysis of locomotion with orientated vector field in a frame. a) The general view of created linear vector field due to robot motion along $z$ axis with velocity $V_v$ b) Kinematic analysis of edges motion with respect to the frame and corresponding object.}
\label{Fig:VectorField}
\end{figure}

The parameters of the LCS filter are divided into two groups i.e., experts and temporary parameters [see Figure \ref{Fig:AlgorithmFunctioningMAP}]. In the temporary parameters, the collector $\pmb{\lambda}$ is responsible for grouping the edges with the location of the center $\uvec{L}_{\lambda}$ and the detection radius size $\mu_{\lambda}$. The ignorance parameter $\pmb{\psi}$ is for $\pmb{\lambda}$ updates in which it takes the information from Circle and Square experts to create ignorance regions via $\uvec{L}_{\psi}$, $R_{\psi}$ and $Ty_{\psi}$ where they are the location of the center, the radius of the region and a flag for geometry type, respectively. The ignorance parameter $\pmb{\psi}$ removes unnecessary edges during the collection of edges by $\pmb{\chi}$. This helps the filter to process faster and concentrate more on the essential locations. Also, $\pmb{\alpha}$ is for detecting certain landmarks that don't follow the main direction of the vector field (along with the robot motion, see Figure \ref{Fig:VectorField} (a) as an example). Next, parameters of the experts carry information from previous frames for each of them. Each expert parameter has a collection series of information about the location $\uvec{L}$, size/radius of the region $R/\mu$, trust factor $Tr$, angle respect to origin $\beta$, the velocity of the corresponding landmark/layer $V$. Also, $\uvec{O}=[O_x,O_y]$ stands for the origin of the landmark/layer of the expert after the first detection. Note that we define layer as a group of landmarks that have similar kinematics such as velocity, the direction of motion, or being in the same region on the frame. These layers not only can stand for a certain region of objects but also create an abstract presentation for better evaluation of the scene by experts. $\uvec{E}_n$ and $\uvec{E}_r$ are the parameters responsible for estimated edges to contain their properties. Besides these parameters in the Circle expert, $\uvec{C}_n$ and $\uvec{C}_r$ are used to assist the system to group the edges with relevant kinematic properties. Finally, $\uvec{S}$ is the Square expert where it keeps the highest level of information about existing objects in the layer form. This final expert matches the objects based on defined geometric and kinematics conditions. 

We have four transitions in this online learning algorithm. Each transition carries detection, learning, and tracking operations. Experts of the filter work in integration with the dynamic learning without sole reliance on a pyramidal supervisory \cite{Bouguet00pyramidalimplementation}. At first, the camera detects the edges and saves it in $\pmb{\chi}$, and the Line Expert groups/removes them by ($\pmb{\lambda}$,$\bm{\psi}$) with feedback from Circle and Square experts. At the transition I, collected edges $\pmb{\chi}$ are transferred to the Circle expert. Circle expert estimates its layers of the landmarks. Next, there is a transition II that updates the ignorance regions $\pmb{\psi}$. Also, the updated Circles (layered landmarks) are transferred to Square expert in transition III. The Square expert geometrically matches the exiting objects in the environment with data from the estimated circles and its predicted data. Finally, the ignorance region $\pmb{\psi}$ for the next edge collection (incoming frame) is updated by the Square expert.

\subsection{Kinematics}
The kinematics of a moving robot is sketched as Figure \ref{Fig:VectorField}, where a holonomic robot with velocity of $V_{\nu}$ travels $D_v$ distance in the direction of $z$ axis.  \makehighlight{In this work, we consider a motion field model rather than a linear rotational models that was used in our previous study \cite{tafrishi2017line}. The rotation is getting updated by the motion field kinematics where the translation is decoupled from the rotation \cite{Collins2007Vision}.
 The location of landmarks/layers $\uvec{L}(k-1)=[L_x,L_y]$ from $k-1$ step including the origin points $\uvec{O}$ (rebel landmarks/layers have their own origin \cite{tafrishi2017line}) are updated by
\begin{eqnarray}
&{L}_x(k-1) &= L'_x(k-1)+O_{I,x}+f\omega_y+{L}'_y(k-1)\omega_z \nonumber\\
&+ &\frac{1}{f}\left[ {L}'_x(k-1){L}'_y(k-1)\omega_x- {L}'^2_x(k-1) \omega_y\right ] \nonumber\\
& {L}_y(k-1) &= L'_y(k-1)+O_{I,y}+f\omega_x+{L}'_x(k-1)\omega_z \nonumber\\
&+& \frac{1}{f}\left[ {L}'_x(k-1){L}'_y(k-1)\omega_y- {L}'^2_y(k-1) \omega_y\right ] \nonumber\\
\end{eqnarray}
 where $\uvec{L}'(k-1)=[L'_x(k-1),L'_y(k-1)]$, $f$ and $\pmb{\omega}=[\omega_x,\omega_y,\omega_z]$ are the location before the rotational update with respect to the frame origin $\uvec{O}_{I}=[O_{I,x},O_{I,y}]$, camera's focal length and angular velocities of the moving vehicle that comes from the IMU. Additionally, the angular direction of the features/layers $\beta(k-1)$ is calculated based on $\uvec{L}(k-1)$ and $\uvec{O}(k-1)$.} It is important to note that we have assumed the rotation changes of the robot (it is independent of depth) with small values in each frame while the decomposed translation motion is estimated using our filter. One can obtain an accurate model for large rotations by utilizing a depth estimation of a Stereo camera \cite{hartley2003multiple} or applying advanced models of the motion field. As the camera moves in an aligned direction, a vector field can be expressed for the normally distributed locomotion of landmarks on the frame [see Figure \ref{Fig:VectorField} (b)]. This linear outward vector field will help us to develop a way to distinguish fixed or low-velocity objects that moves along the field from the abnormal landmarks, namely "rebel" landmarks/layers. These rebel landmarks can be an object that comes toward the camera or be an independently moving object that does not follow the vector field of the vehicle motion. 

\subsection{Trust Factor}
\begin{figure}
\centering
\includegraphics[width=3.2 in]{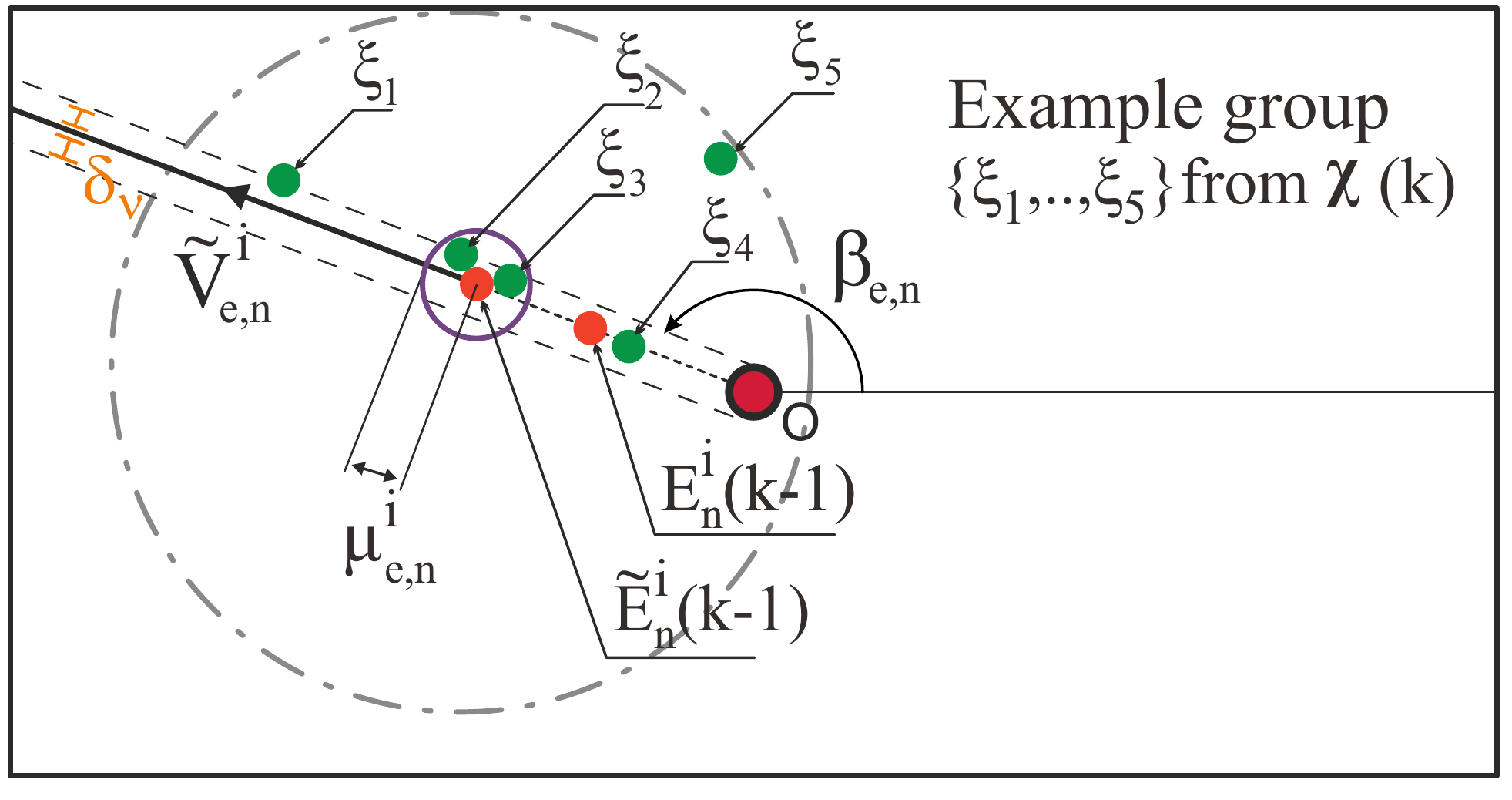}
\caption{The edge's classifications with current information of $\chi$.}
\label{Fig:EdgeEvaluationforlocationesti}
\end{figure}
The trust factor as an important part of the LCS filter is a variable that evaluates the reliability of the detected edge or layers. The trust variable $ Tr \;\in \; N $ is the main factor in the learning part of the filter that is parameters with three ranks:

\textbf{$Tr_s$}:
The standard trust means the trust factor $Tr$ with equal and greater than the value of $Tr_s$ is landmarks/layers that are highly accurate with minor errors. Also, a high accuracy here means kinematic approximation matches with collected data. Thus, the trust factor of the parameters that reaches this value is possible to be tracked about their accuracy level respect to existing data from the landmark/layer. 

\textbf{$Tr_{c}$}: If the trust factor of edges/layers $Tr$ hit the values less than the critical value $Tr_{c}$, they get deleted. A decrease in trust value to below $Tr_c$ means these landmarks belong to objects that already passed or the wrong estimation happened in the frame.

\textbf{$Tr_{m}$}: The maximum value in trust $Tr$ that limits the parameters in the filter to a certain value for preventing overconfidence. Also, it helps in evaluations of the ignorance parameter $\pmb{\psi}$.

Each of these ranked trusts lets the filter properly eliminate, re-coordinate, or combine the edges and layers of the Line, Circle and Square experts. Note that every newly created normal landmarks/layers i.e., normal edges $\uvec{E}_n$ and $\uvec{C}_n$ circles, have their initial trust factor as $Tr(0)=\frac{1}{2}[Tr_{c}+Tr_{s}]$. A new candidate in the Square expert $\uvec{S}$ has the initial $Tr(0)=Tr_{s}$ trust factor. Also, because tracking the rebel landmarks/layers i.e., $\uvec{E}_r$ and $\uvec{C}_r$, are harder, they will have an initial trust factor larger than $Tr(0)>Tr_{s}$ in contrast to normal ones. 

%% file: contents/lcs.tex
\section{Experts of the LCS Filter} \label{sec:lcs-filter}

\subsection{Line Expert}
The Line expert evaluates the landmarks that are collected from raw images of camera. This expert has two tasks: the first is removing unnecessary/repetitive information; The second is grouping the landmarks for faster evaluations in the next expert. 

\begin{figure}
	\centering
	\includegraphics[width=2.5 in]{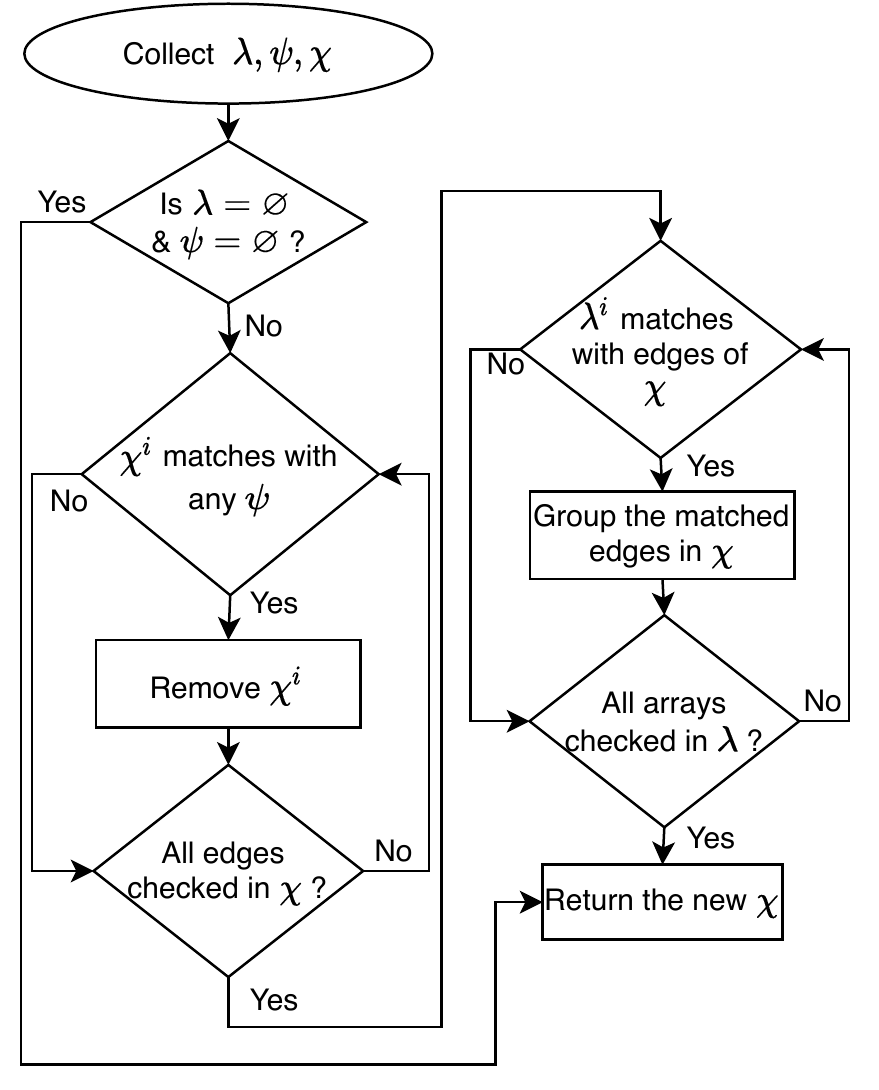}
	\caption{Flowchart of the Line expert.} 
	\label{Fig:LineExpert}
\end{figure}

The flowchart in Figure \ref{Fig:LineExpert} shows the computation of the Line expert. 
At first, the landmarks are reduced in $\pmb{\chi}$ with using the feed of ignorance regions $\pmb{\psi}$ since the high-level experts (the Circle and Square) grantee that the landmarks in $\pmb{\psi}$ region are unnecessary to be updated for certain frame numbers. We will explain how the trust factor $Tr$ will help the expert evaluation to predict certain regions with decreasing computations in the next steps. Next, $\pmb{\lambda}$ is grouping the landmarks in $\pmb{\chi}$ with given variables ($L_{\lambda},\mu_{\lambda}$) in each frame. Also, newly appeared unmatched edges with $\pmb{\lambda}$ are added as a single landmark in $\pmb{\chi}$. \makehighlight{ Grouping temporary parameter $\pmb{\lambda}$ detects candidate edges in $\pmb{\chi}$ by their distance $\parallel \uvec{L}_{\lambda} - \uvec{L}_{\chi}\parallel$ from distance radius $\mu_{\lambda}$ \cite{tafrishi2017line}. }
Next, the ignorance parameter $\pmb{
\psi}$ removes the edges via including the expressed regional constraints for different geometric flag $Ty$ as 
\begin{equation*}
\begin{cases}
&\pi R_{\psi}^2,\;\;\;\;\;(L_{\psi,x},L_{\psi,y})\;\;\;\;\;\;\;\;\;\;\;\;\;\;\;\;\; Ty\;=\;1\\
& R_{\psi,x}\cdot R_{\psi,y},\;\;\;\;\;(L_{\psi,x},L_{\psi,y})\;\;\;\;\;\;\; Ty\;=\;2
\end{cases}
\label{Psijun}
\end{equation*}
where $Ty=1$ and $Ty=2$ flags are presenting the circular and rectangle areas. Finally, the flowchart in Figure~\ref{Fig:LineExpert} returns new grouped $\pmb{\chi}$. 
\subsection{Circle Expert}
The Circle expert combines the landmarks with kinematic comparison at the grouped data. This stage transforms individual landmarks into grouped dynamic patterns. This expert increases the level of information to determine partially recognized layers in the scene by the next expert.

\subsubsection{Edge Classifications}
The Circle expert classifies the detected landmarks $\pmb{\chi}(k)$ in the direction of the moving robot that creates the vector field (normal edges) with respect to their predicted $\tilde{\uvec{E}}_n(k-1)$ landmarks as shown in Figure \ref{Fig:EdgeEvaluationforlocationesti}. By considering the center of the captured image aligned with $z$ axis of the robot motion as Figure \ref{Fig:VectorField}, the detected landmarks are classified with five basic scenarios $\{\xi_1,...,\xi_5\}$ depending on their location with respect to predicted landmarks $\tilde{\uvec{E}}_n(k-1)$ on the image. Note that the properties of $V_{e,n}$ and $\beta_{e,n}$ (edge angle) are for the edges that are mostly passed objects as called ``normal edges''. This is why the velocity is aligned with the center of frame $\uvec{O}_{I}$. The motion flow of normal edges vectors is linear since we assume robot holonomic motion is always aligned with $z$ axis. However, the center for the rebel edges $\uvec{O}_{e,r}$ are different depending on the coming origin from $\pmb{\alpha}$. The dashed double lines express the angular error span $\delta_{\nu}$ that is dependent on the model error and vehicle velocity (obtained by IMU sensor). Also, the normal edge boundary radius $\mu_{e,n}$ is getting updated proportional to the numbers of obtained landmarks and their kinematics.

Before giving the prediction definitions, the trust factor changes in each frame depends on the obtained evaluations through proposed classifications presented in Fig \ref{Fig:EdgeEvaluationforlocationesti}. This tells us how far the approximated landmarks are reliable for estimations by the captured edges. Not only these classifications let us determine new and low-accuracy landmarks but also filter can recognize rebel and normal moving edges from each other. Assuming the $i$-th landmark candidate in $\uvec{E}^i_{e,n}$, the estimations are classified in five cases:
\begin{figure}
	\centering
	\includegraphics[width=\columnwidth]{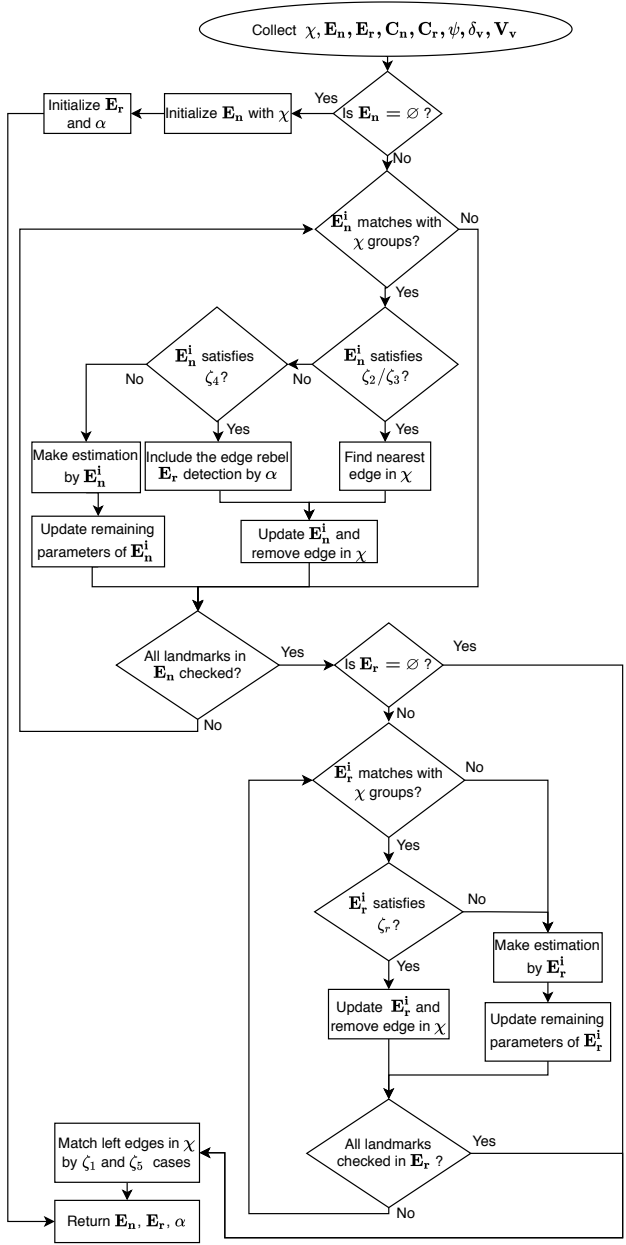}
	\caption{Flowchart of the Circle Expert for edge estimation.} 
	\label{Fig:EdgeEstimation2} 
\end{figure}
\begin{figure*}
	\centering
	\includegraphics[width=2 in]{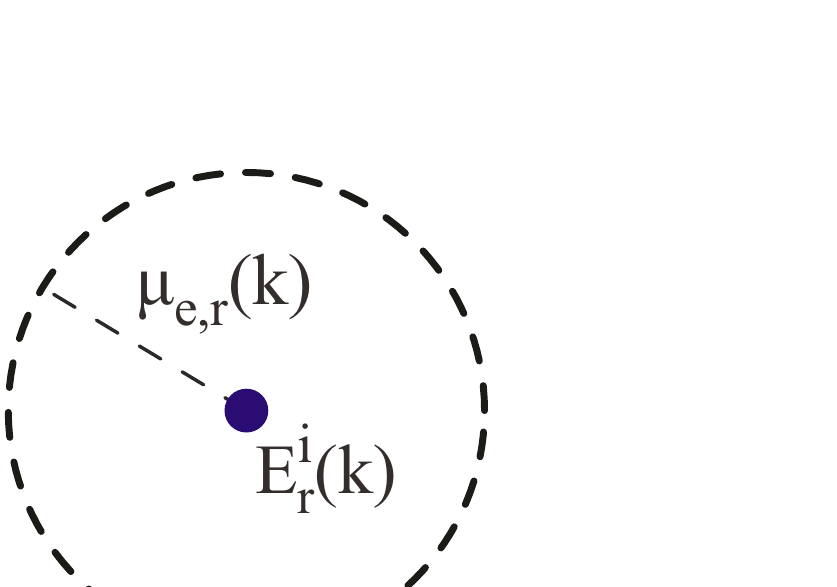}
	\includegraphics[width=2 in]{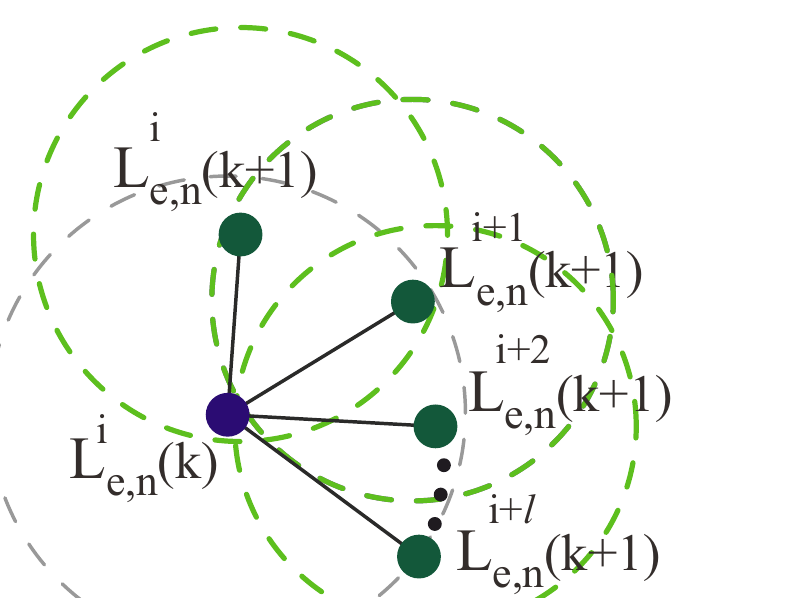}
	\includegraphics[width=2 in]{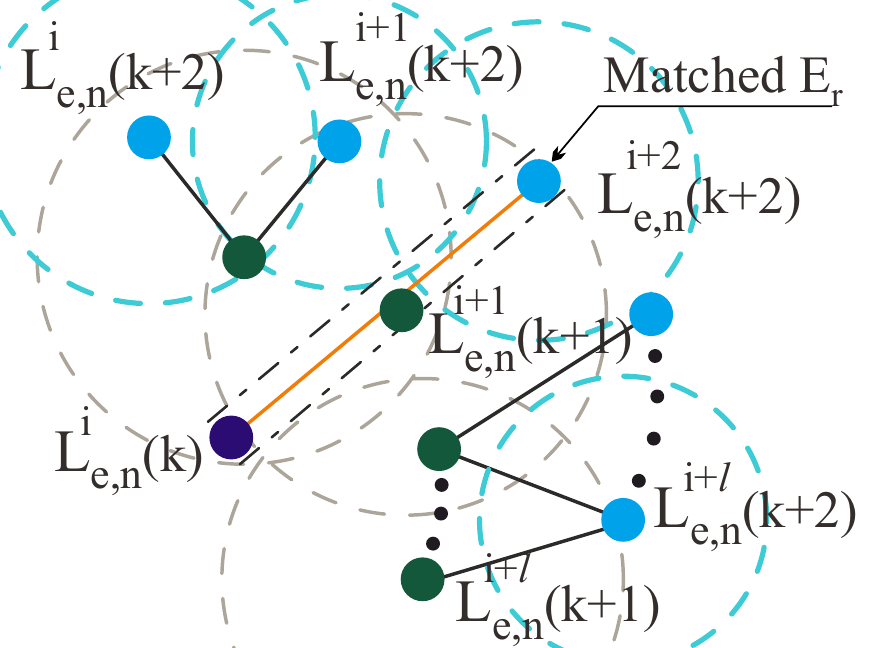}
	\newline
	\space\space\space\space\space\space\space\space\space\space\space\space\space\space\space\space\space\space\space\space\space\space  Frame k\space\space\space\space\space\space\space\space\space\space\space\space\space\space\space\space\space\space\space\space\space\space\space\space\space\space\space\space\space\space\space\space\space\space\space\space\space Frame k+1 \space\space\space\space\space\space\space\space\space\space\space\space\space\space\space\space\space\space\space\space\space\space\space\space\space\space\space\space\space\space\space\space\space\space\space\space\space\space Frame k+2 \space\space\space\space\space\space\space\space\space\space\space\space\space\space\space\space\space\space\space\space\space 
	\caption{Rebel landmarks detection through within the three frames. The dashed gray lines are the previous step radius size, the dot-dashed lines are the error span variation's space $\delta_v$. The remaining colorized dashed circles are corresponding $\mu_{e,r}$ radius size.}
	\label{Fig:EDgerebellers}
\end{figure*}

$\bullet$ \textbf{$\xi_1$}: This detected landmark in $\pmb{\chi}$ is out of the $\mu_{e,n}$ radius of the predicted edge $\tilde{\uvec{E}}^{i}_n(k-1)$ but exists in the error span $\delta_{\nu}$. This detection will be added as an extra normal edge inside the included circle but the trust value $Tr$ of estimated edge $\hat{\uvec{E}}^{i}_n(k)$ will be reduced by 1.

$\bullet$ \textbf{$\xi_2$}: This edge is successfully within the area of $\mu_{e,n}$ and the error span $\delta_{\nu}$; Hence, they are matched for estimation with approximated edge $\tilde{\uvec{E}}^{i}_n(k-1)$. The trust $Tr$ of the corresponding edge is added with 1.

$\bullet$ \textbf{$\xi_3$}: This edge is in the area of $\mu_{e,n}$ but it fails from approximated location $\tilde{\uvec{L}}_{e,n}(k-1)$ and velocity $\tilde{V}_{e,n}(k-1)$ of the chosen edge $\uvec{E}^i_{e,n}(k-1)$. To include the approximation error for incoming estimation, the trust value $Tr$ for estimated edge will be reduced by 1. 

$\bullet$ \textbf{$\xi_4$}: Despite the exclusion from approximated edge radius $\mu_{e,n}$, it is in the error span of $\delta_{v}$. Thus, we call it a candidate for being a rebel landmark. Thus, the trust of estimated normal edge $\uvec{E}^i_{e,n}(k-1)$ will be decreased by 1 but the information will be carried to $\pmb\alpha$ for the possible existence of rebel edges $\uvec{E}_r$.

$\bullet$ \textbf{$\xi_5$}: This case fails all the predictions of velocity $\tilde{V}_{e,n}(k-1)$ and location $\tilde{\uvec{L}}_{e,n}(k-1)$ as well as regional constraints $\mu_{e,n}$ and $\delta_v$; Hence, it is added as a rebel edge candidate at $\pmb{\alpha}$ or new normal edge based on their failed categories. The trust value of estimated edge $\uvec{E}^i_{e,n}(k-1)$ will decrease by 1. \makehighlight{In here, the condition for satisfying the error span $\delta_{v}$ is defined by the similar formula in our previous study \cite{tafrishi2017line} respect to the image center $\uvec{O}_I$ and the detected landmark location $\uvec{L}_{\chi}$.} 

The flowchart in Figure \ref{Fig:EdgeEstimation2} shows how the Circle expert evaluates and detects the edges. At the initiation of $\uvec{E}_n$ matrix, landmarks velocities are equal to the vehicle velocity $V_{e,n}(0)=V_v$, the angle $\beta_{e,n}(0)$ is defined depending on the edge location $\uvec{L}_{e,n}(0)$ respected to frame center $\uvec{O}_I$. Also, the radius of detection $\mu_{e,n}(0)$ is considered with a constant $\mu_0$ value where it will be updated relative to the grouped landmarks $\pmb{\chi}$ later on. We define the estimations of certain variables e.g., $\uvec{L}$, $\beta$ and $R$, in Figure \ref{Fig:AlgorithmFunctioningMAP} at $k$-th frame as follows
\begin{align}
    \hat{\uvec{P}}(k)=\frac{(Tr(k-1)-Tr_{c})\uvec{P}'+\uvec{P}(k)}{(Tr(k-1)-Tr_{c})+1}
    \label{Eq:estimationgeneralfor}
\end{align}
where $\uvec{P}(k)$ and $\uvec{P}'$ are two arbitrary vectors. For example, $i$-th normal edge in $\uvec{E}_n$ at $k$-th frame is estimated for its location $\uvec{L}_{e,n}$, velocity $V_{e,n}$ and detection radius $\mu_{e,n}$ and the angular direction $\beta_{e,n}$ as follows
\begin{align}
\begin{split}
&\hat{\uvec{L}}^i_{e,n}(k)=\frac{(Tr_{e,n}(k-1)-Tr_{c})\tilde{\uvec{L}}_{e,n}(k-1)+\uvec{L}_{\chi}(k)}{(Tr_{e,n}(k-1)-Tr_{c})+1},\\
&\hat{V}^i_{e,n}(k)=\Bigg|V_{v}(k)\pm\frac{\parallel \tilde{\uvec{L}}_{e,n}(k-1)- \uvec{L}_{\chi}(k) \parallel }{t_f}\Bigg|,\;\;\\
&\hat{\mu}^i_{e,n}(k)=\frac{1}{2}\Bigg[\frac{\left|V_{v}(k)-\tilde{V}_{e,n}(k-1)\right|}{Cor(\sum \chi)}+\mu_{e,n}(k-1)\Bigg],\\
&\hat{\beta}^i_{e,n}(k)=\angle \left(\uvec{L}^i_{e,n}(k),\; \uvec{O}_{I} \right)
\end{split}
\label{Eq:EnEstimation}
\end{align}
where $t_f$, $Cor( \cdot )$, $\sum \chi $ and $\angle(\cdot,\cdot)$ are the sampling time, the correlation operation, group of captured edges in $\pmb{\chi}$ matched with the predicted landmark $\tilde{\uvec{E}}^i_{e,n}(k-1)$ and the angle of point $\uvec{L}^i_{e,n}(k)$ with respect to frame origin $\uvec{O}_I$, in the given order. It is assumed that the robot recovers the angular error $\delta_{v}$, due to camera vibrations and IMU estimation errors,
in each step. Thus, the motion angle of the edge $\beta_{e,n}$ is kept the same. However, $\beta_{e,n}$ should be predicted in early steps to find the corresponding edge classification [see Figure \ref{Fig:EdgeEvaluationforlocationesti}] where the predicted velocity is $\tilde{V}_{e,n}=\frac{1}{2}[V_{e,n}(k-1)+V_{\nu}]$.

For detecting the rebel edges $\uvec{E}_r$, we propose a line tracking model that tries to find these edges within N-frame steps [see Figure \ref{Fig:EDgerebellers}]. By presenting the minimum $N=3$ frame, we include the landmarks into the matrix of the rebel landmarks alignment $\pmb{\alpha}$ that satisfy $\zeta_4$ and $\zeta_5$ classifications within the $N$ frame. We apply the detection with considering the underneath specifications:

$\bullet$ Due to unexpected motion in the rebellious landmarks for consecutive frames, tracking will be a  non-linear motion with a certain deviation in each frame.

$\bullet$ The number of frames to evaluate the reliability of existing detection to rebel landmarks will be $N=3$.

$\bullet$ The relation of connected landmarks is deleted after success/fail in every $N$ frame analysis of $\pmb\alpha$. 

\begin{figure}
	\centering
	\includegraphics[width=3 in]{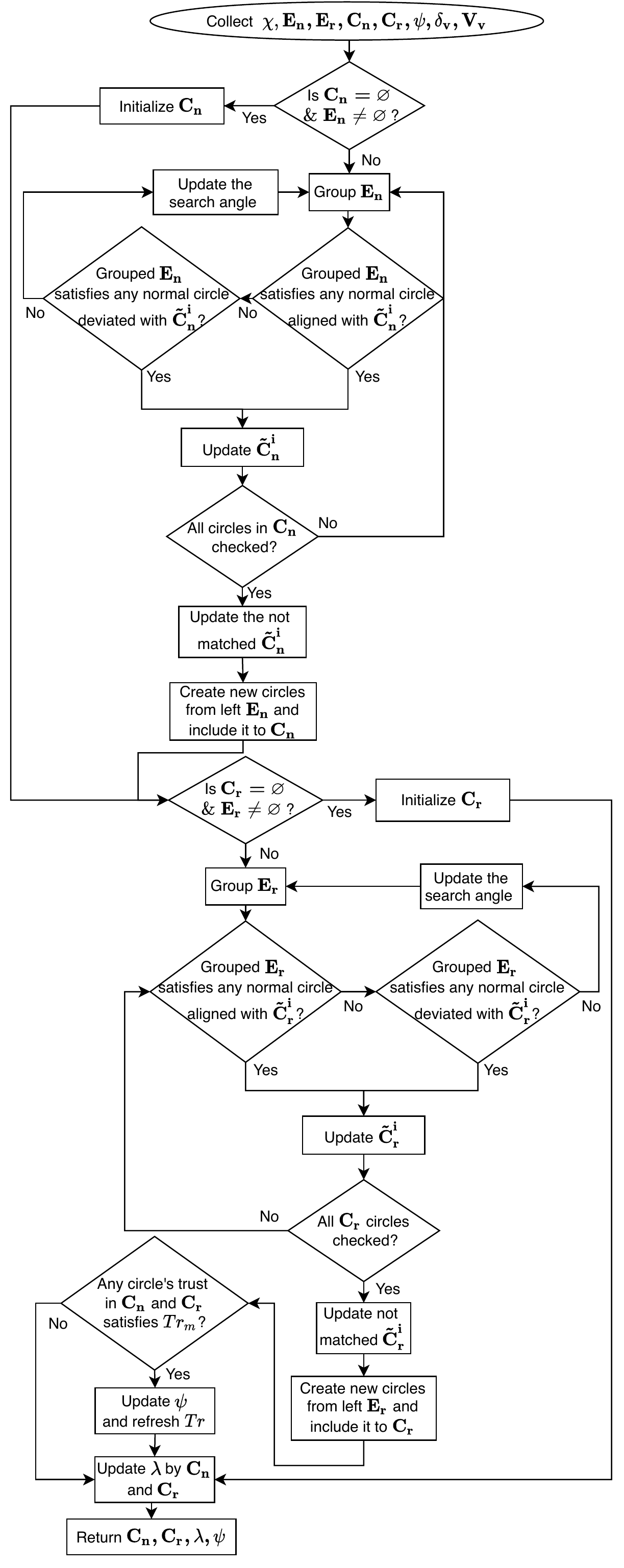}
	\caption{\makehighlight{Flowchart of the Circle expert for circle estimation part.}}
	\label{Fig:CircleEstimation}
\end{figure}

The matrix of rebel landmarks alignment $\pmb\alpha$ collects the locations of edges in the following way 
\begin{equation}
\pmb{\alpha} = \left[\begin{array}{ccc}
\uvec{L}^i_{e,n}(k)  &\uvec{L}^{i}_{e,n}(k+1)  &\uvec{L}^{i}_{e,n}(k+2) \\
\uvec{L}^i_{e,n}(k) &\uvec{L}^{i}_{e,n}(k+1) &\uvec{L}^{i+1}_{e,n}(k+2)\\
\uvec{L}^i_{e,n}(k) & \uvec{L}^{i+1}_{e,n}(k+1)& \uvec{L}^{i+2}_{e,n}(k+2)\\
\vdots & \vdots & \vdots\\
\uvec{L}^i_{e,n}(k) & \uvec{L}^{i+l}_{e,n}(k+1)& \uvec{L}^{i+l}_{e,n}(k+2)\\
\end{array}\right]
\end{equation}
where, as an example, $\uvec{L}^i_{e,n}(k)$ is the location of $i$-th edge candidate at $k$-th frame. In this case, Figure \ref{Fig:EDgerebellers} shows the successfully matched rebel edge that is obtained in the third row of the collection. Note that failed candidates get removed from matrix $\pmb\alpha$ every frame. The initial values for constructed rebel edges at $\uvec{E}_r$ are defined by
\begin{eqnarray}    
\uvec{O}_{e,r}(0)&=&\uvec{L}^i_{e,n}(k),\nonumber\\
{V_{e,r}}(0)&=&\frac{\parallel \uvec{L}^{i+2}_{e,n}(k+2)-\uvec{L}^{i+1}_{e,n}(k+1) \parallel}{t_f}, \nonumber\\
\beta_{e,r}(0)&=& \angle \left(\uvec{L}^{i+2}_{e,n}(k+2),\; \uvec{L}^{i}_{e,n}(k)\right),\nonumber \\
\mu_{e,r}(0)&=&   \angle \left(\uvec{L}^{i+2}_{e,n}(k+2),\; \uvec{L}^{i}_{e,n}(k)\right) \nonumber\\
&- & \angle \left(\uvec{L}^{i+1}_{e,n}(k+1), \;\uvec{L}^{i}_{e,n}(k)\right)
\label{Eq:Intialrebelpara}
\end{eqnarray}
where $\mu_{e,r}$ is defined as the deviation angle here which is different from the detection radius of the normal edge $\mu_{e,n}$. \makehighlight{ These rebel landmarks do not follow the normal vector field of the vehicle motion; Hence, their kinematics are defined by detected location information in $\pmb{\alpha}$}. For the landmarks that already exist in the rebel edge matrix $\uvec{E}_r$, their location $\uvec{L}_{e,r}$ and ${V}_{e,r}$ velocity are estimated with same equations in equations~(\ref{Eq:estimationgeneralfor}) and (\ref{Eq:EnEstimation}), respectively. \makehighlight{ However, the deviation angle of $\mu_{e,r}$ is updated by $\hat{\mu}_{e,r}(k)=|\tilde{\beta}_{e,r}(k)-\hat{\beta}_{e,r}(k-1)|$,
where the predicted angle $\tilde{\beta}_{e,r}(k)$ is calculated from the last detected edge respected to rebel edge's origin $\uvec{O}_{e,r}$ \cite{tafrishi2017line}.} These kinematic formulations make the detection and learning to be smooth and have a similar appliance of negative and positive classified examples \cite{TLD2012} in here. In contrast,  gradual learning is used with a disappearing pattern via the trust factor definitions. Note that $\zeta_r$ classification for the rebel edges is similar to $\zeta_2$ while the detected edge is inside both $\delta_v$ and $\mu_{e,r}$ constraints.

The computation flow of the given flowchart in Figure \ref{Fig:CircleEstimation} is to  prepare the normal and rebel circles as the layers of the detected objects in the frame.
This designed expert is divided into three main parts. The first part was about detecting and comparing the collected group edges $\pmb{\chi}$ with $\uvec{E}_n$ as explained. Next, after elimination, remaining unmatched landmarks are carried to the second part for the rebel edges $\uvec{E}_r$ study which was based on flowchart in Figure~\ref{Fig:EdgeEstimation2}. Then, $\uvec{E}_n$ and $\uvec{E}_r$ matrices are compared under certain kinematic comparisons to develop the latest circles as layers [as shown in Figure \ref{Fig:CircleEstimation}].
\subsubsection{Normal Edge Circling}
In the normal circles, collected landmarks are following the direction of the vector field [see Figure \ref{Fig:VectorField}] and they are categorized depending on their directions of the motion, velocities and, locations on the image.
\begin{figure}
\centering
\includegraphics[width=0.9\columnwidth]{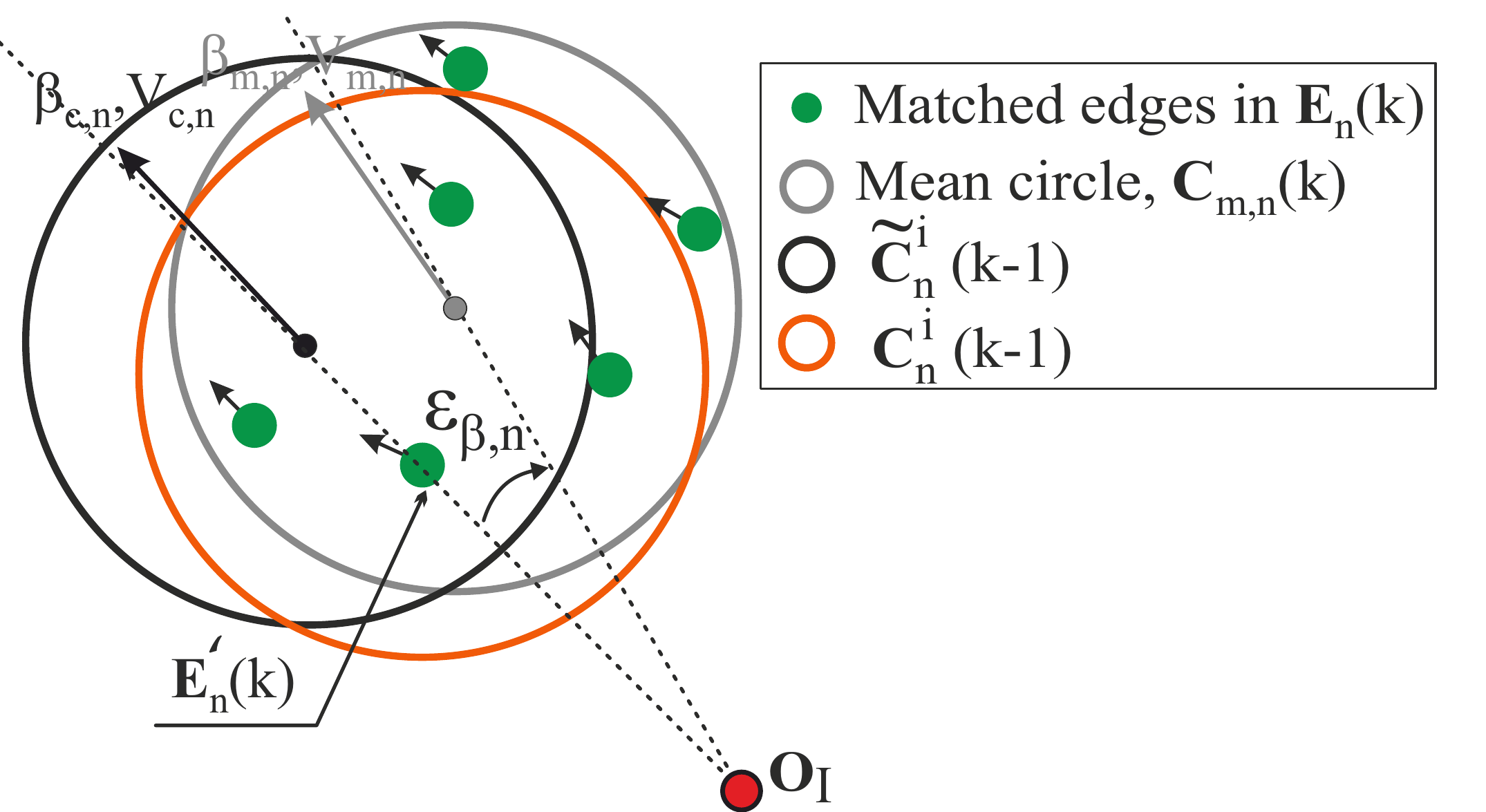}
\caption{Circle matching in normal landmarks.}
\label{Fig:Normaledge1}
\end{figure}

Figure \ref{Fig:Normaledge1} shows how the edges with the alike kinematics are matched in the expert. In this matching, the estimated circle ${\small\hat{\uvec{C}}^i_{n}(k)}$ in the $k$-th frame is determined by previously predicted normal circle ${\small \tilde{\uvec{C}}^i_{n}(k-1)}$ of the frame $k-1$ and the grouped new circle $\uvec{C}_{n,m}(k)$ with estimated edges in the $k$-th frame. At first, the chosen candidate normal edge $\uvec{E}'_n$ is matched with other landmarks depending on their angle $\beta'_{e,n}$ and velocity $V'_{e,n}$ to construct $\uvec{C}_{n,m}(k)$ as follows
\begin{equation}
\begin{cases}
&\left (\beta'_{e,n} - \varepsilon_{\beta,n} \right )< \beta^i_{e,n} < \left ( \beta'_{e,n} + \varepsilon_{\beta,n} \right ) \\
& |V'_{e,n}| \leq |V^i_{e,n}|+\varepsilon_{v,n} V_{v} 
\end{cases}
\;\;\;\;\;\;\;\;\;\;\;\;\;1<i\leq l
\label{EdgeEncompare}
\end{equation}
where $\varepsilon_{\beta,n}$ and $\varepsilon_{v,n}$ are the accuracy constants of angle and velocity for the normal circle. Next, to compare the predicted circle ${\small\tilde{\uvec{C}}^i_n(k-1)}$ with the mean circle of the grouped edges $\uvec{C}_{m,n}$, the following condition is used:
\begin{equation}
\left[(\beta_{m,n} - \varepsilon_{\beta,n}) < \beta^i_{{c,n}}  <( \beta_{m,n} + \varepsilon_{\beta,n}) \right]\; \& \;
 \left[ V_{m,n} \leq \varepsilon_{v} V^i_{c,n} \right]
\label{Eq:EdgeCirclenormalcomparee}
\end{equation}
where 
\begin{align*}
  \beta_{m,n}=  \frac{1}{l} \sum^{l}_{q=1} \beta^q_{e,n}, \;  V_{m,n}=  \frac{1}{l} \sum^{l}_{q=1} |V^q_{e,n}|
\end{align*}

Together with matching conditions in (\ref{Eq:EdgeCirclenormalcomparee}), there is a weighting function that finds out the percentage of the overlap. The percentage of overlap finds the ratio of the located edges $q\in[1,l]$ in the mean circle of the grouped edges $\uvec{C}_{m,n}$ are in the predicted circle ${\small\tilde{\uvec{C}}^i_n(k-1)}$ as
\begin{equation}
Per \left[  \parallel   \uvec{L}^q_{e,n}(k)-\tilde{\uvec{L}}_{c,n}(k-1) \parallel  < R^q_{c,n}(k-1) \right] < \rho_c
\label{Eq:percentinvolve}
\end{equation}
where $Per[\cdot]$ and $\rho_c$ are the percentage operation that calculates the ratio of edges in $\uvec{C}_{m,n}$ which are located inside of ${\small\tilde{\uvec{C}}^i_n}$ and the minimum overlap percentage, respectively.

The estimated circles are updated with certain similar functions as landmarks. For example, the location $\hat{\uvec{L}}_{c,n}(k)$, the radius $\hat{R}_{c,n}(k)$ and direction angle $\hat{\beta}_{c,n}(k)$ are estimated by equation~(\ref{Eq:estimationgeneralfor}). 
While the predicted circle's angle $\tilde{\beta}^i_{c,n}(k-1)$ is aligned with averaged group of edges in $\uvec{C}_{m,n}$ within the accuracy constant $\varepsilon_{\beta,n}$, the estimated circle ${ \small \hat{\uvec{C}}^i_{c,n}(k)}$ is upgraded with +1 trust factor. If the angle has deviation with respect to the estimated one despite major inclusion of edges in the circle, the trust factor is updated with -1. 
\subsubsection{Rebel Edge Circling}
\begin{figure}
	\centering
	\includegraphics[width=0.9\columnwidth]{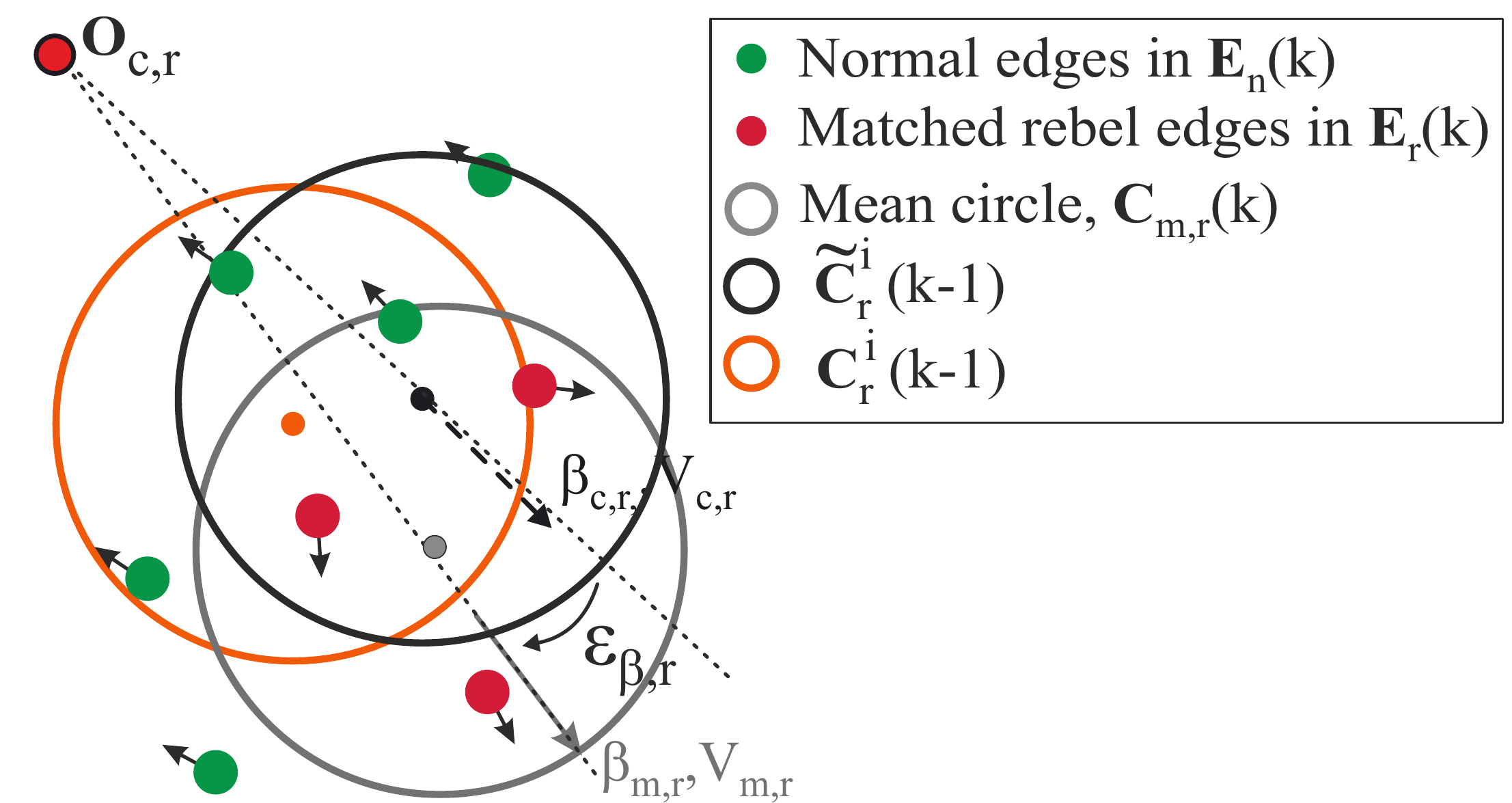}
	\caption{Circle matching in the rebel landmarks.}
	\label{Fig:Normaledge2}
\end{figure}

Certain objects in the scene will have inconsistent motion in the frame. These objects normally will be the ones that have potential to come toward the robot or are moving objects in an arbitrary directions (such as side passing cars). Therefore, detected rebellious edges will have the most importance when it comes to SLAM or object avoidance applications which is one of the strength of the proposed filter. The rebel landmarks in the worst-case scenario can be appeared within the normal edge circles [see Figure \ref{Fig:Normaledge2}]. \makehighlight{ To detect the rebel edges, previously available similar edges are considered with specific dedication to the $\beta_{c,r}$ and $V_{c,r}$ similar to condition (\ref{EdgeEncompare}) 
where it has accuracy constants of the angle $\varepsilon_{\beta,r}$ and $\varepsilon_{v,r}$ velocity in grouping rebel edges. }It must be said that these edges are the hardest ones since they are not following vectors field flow ($\varepsilon_{v,r}>\varepsilon_{v,n}$ and $\varepsilon_{\beta,r}>\varepsilon_{\beta,n}$) when they are having a nonlinear displacement (unpredictable motion) in frame.

Based on the characteristics of the rebel edges, the velocity condition is largely dependent on rebel edges estimated velocities and the mean circle of grouped rebel edges. Thus, a condition is developed for detecting the right grouped rebel edges $\uvec{C}_{m,r}(k)$ with the predicted $i$-th rebel circle ${\small \tilde{\uvec{C}}^{i}_r(k-1)}$ as follows
\begin{equation}
\begin{split}
&\left[ \left ( \beta_{m,r} - \varepsilon_{\beta,r} \right) < \beta^i_{c,r}  < \left ( \beta_{m,r}  + \varepsilon_{\beta,r} \right ) \right]  \; \\  & \& \; \left[V_{m,r} \leq( V^i_{c,r}+\varepsilon_{v,r} V_{\nu}) \right]
\end{split}
\label{Eq:EdgeCirclerebelcompareER}
\end{equation}
where
\begin{align*}
  \beta_{m,r}=  \frac{1}{l} \sum^{l}_{q=1} \beta^q_{e,r}+\mu^q_{e,r}, \;  V_{m,r}=  \frac{1}{l} \sum^{l}_{q=1} |V^q_{e,r}|
\end{align*}
The same overlap condition (\ref{Eq:percentinvolve}) for locating rebel circles are taken place as normal circles. The location $\hat{\uvec{L}}_{c,r}(k)$ and radius $\hat{R}_{c,r}(k)$ and trust factor $Tr_{c,r}$ are updated similar to the normal circle except in the estimation of angle $\hat{\beta}_{c,r}(k)$, we have
\begin{equation*}
\hat{\beta}_{c,r}= \frac{1}{l}\overset{l}{\sum} [\beta_{e,r}+\mu_{e,r}]
\end{equation*}
Note that overall filter is constructed with the same analyzed flow as Figure~\ref{Fig:CircleEstimation}.


\subsection{Square Expert}
\begin{figure}
    \centering
    \includegraphics[width=.92\columnwidth]{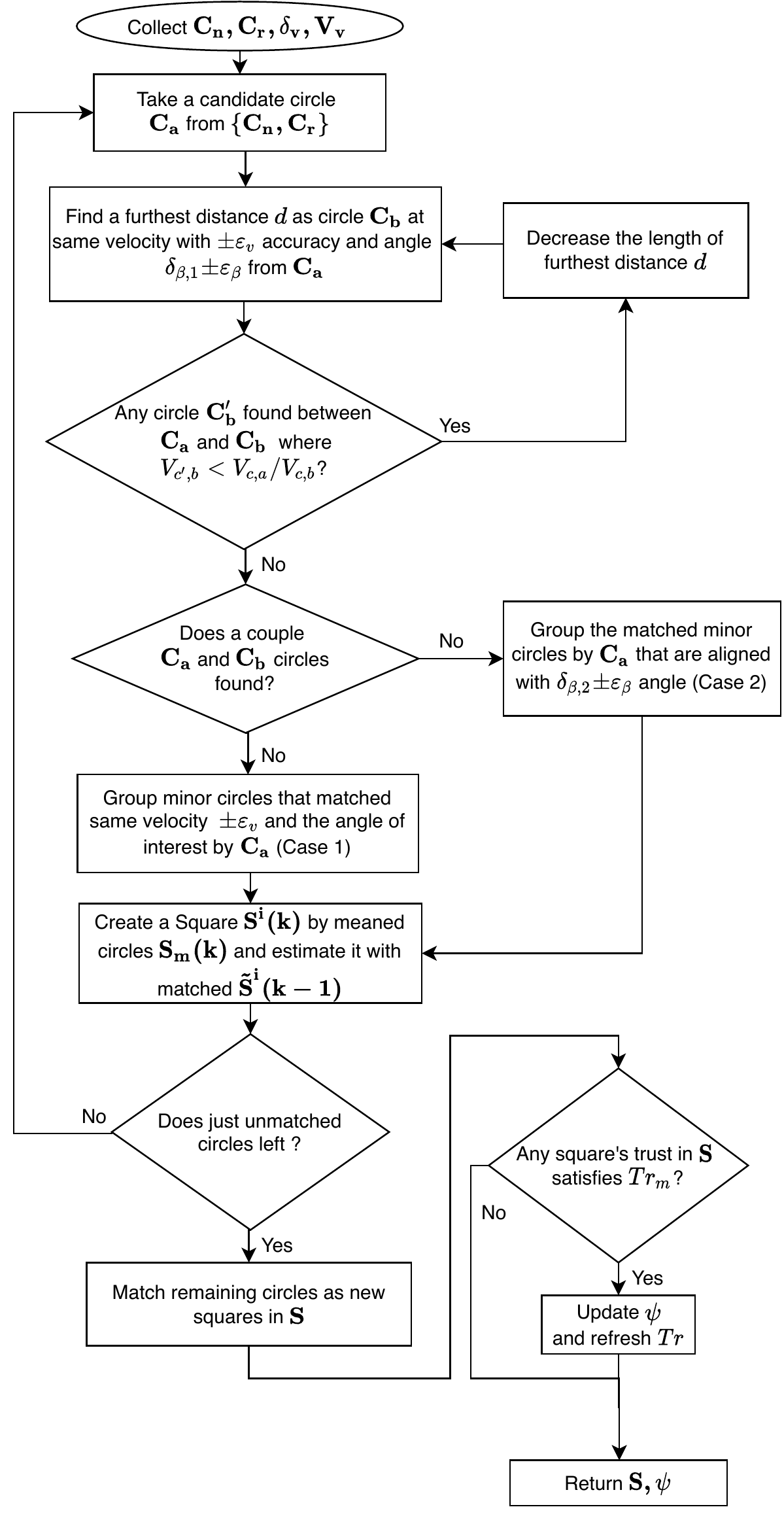}
    \caption{Flowchart of the Square Expert. }
    \label{fig:my_label_SFilter}
\end{figure}
The Square expert geometrically combines circles from the Circle expert to find physical layers. These layers stand for the objects' regions or the object itself. Note that our filter does not contain any recognition source for different objects; Hence, the objects can be packed in layers due to equal distances. This grouping happens by kinematic properties that are captured by the inertial sensors mounted on the moving vehicle. This grants fast evaluation of the incoming scene to group each region with corresponding geometry where it can be either passing or incoming object toward the camera. Note that we here mention each candidate as a "Square" for simplicity but the detected layers can geometrically be any geometry of the rectangle form.  

The flow chart in Figure \ref{fig:my_label_SFilter} shows the workflow of the Square expert. The Square expert mainly consists of two primary parts: in the first part, it collects the circles with their certain geometries and velocity properties. In the second part, this expert matches the grouped circles with the previous step's predicted square. In this expert, after combining the rebel and normal circles, the algorithm tries to find the furthest couple of circles that can correspond to a complete object or part of it. Then, it will cancel out the couple and update the distance if there exist other objects that do not relate to the same layer (distance) of it. The grouped circles (minor circles) with included couples are then transformed into a rectangle geometry $\uvec{S}_m$. Then, the mean square candidate $\uvec{S}_m$ is estimated with the predicted $i$-th square $\tilde{\uvec{S}}^i(k-1)$ in $k-1$ frame. The process continues till all the circles are looped in the frame $k$. 
\subsubsection{Geometric Layering}
In this procedure, a circle is firstly selected as $\uvec{C}_a$ from $\{\uvec{C}_n,\uvec{C}_r\}$, and then the algorithm searches for two cases of geometric matches. The first case (Case 1) is when two circles are matched with $\pm 90^o$ angle from each other [see Figure \ref{fig:my_label_SFilter} as an example]. Case 2 matches circles that are approximately aligned in the same direction and with the same velocity. Note that, here, we choose simple extending square corners and aligned groups matching as layers in the frame. One can make more complex structures, for example, hexagon or smooth closed convex graphs which we think is a separate study in itself.

In case 1, the selected initial circle $\uvec{C}_a$ is matched with its couple circle $\uvec{C}_b$ with the following velocity ($V_{c,a},V_{c,b}$), angle ($\beta_{c,a},\beta_{c,b}$) and distance ($d$) conditions:
\begin{equation}
\begin{split}
&\Big[\left[V_{c,b}-\varepsilon_v\leq V_{c,a}\leq V_{c,b}+\varepsilon_v \right]\; \\
&\& \;
\big[ \beta_{c,b}+\delta_{\beta,1}-\varepsilon_{\beta}\leq \beta_{c,a}\leq \beta_{c,b}+\delta_{\beta,1}+\varepsilon_{\beta}\big] \; \& \;\left[ d<d_{t}\right]\; \Big]\\
&||\;\Big[ \left[V_{c,b}-\varepsilon_v\leq V_{c,a}\leq V_{c,b}+\varepsilon_v \right]\\
&\; \& \;
\big[ \beta_{c,b}-\delta_{\beta,1}-\varepsilon_{\beta}\leq \beta_{c,a}\leq \beta_{c,b}-\delta_{\beta,1}+\varepsilon_{\beta}\big] \; \& \;\left[ d<d_{t}\right] \Big]
\end{split}
\label{Eq:ConditionCoupleCirclematch}
\end{equation}
where $\delta_{\beta,1}$, $d$ and $d_{t}$ are the angle difference for the candidate corner circle $\uvec{C}_b$ in the first case, the distance between couple of circles and the maximum temporary distance between circles $\{\uvec{C}_a,\uvec{C}_b\}$. Initial term in condition (\ref{Eq:ConditionCoupleCirclematch}) matches the similar velocities between the circles. Next, the angle of the counterpart circle $\uvec{C}_b$ should be satisfied which we consider as $\pm 90^o$ here. As mentioned before, more complicated geometries can be assumed but this will complicate the iterations due to the grouping stage for other circles. Next, the distance of the two circles  $d=||\uvec{L}_{c,a}-\uvec{L}_{c,b}||$ has to be always smaller than the maximum temporary distance $d_{t}$ where $\uvec{L}_{c,a}$ and $\uvec{L}_{c,b}$ are the locations of circle candidates $\uvec{C}_a$ and $\uvec{C}_b$, respectively.

\begin{figure}
    \centering
    \includegraphics[width=2.8 in]{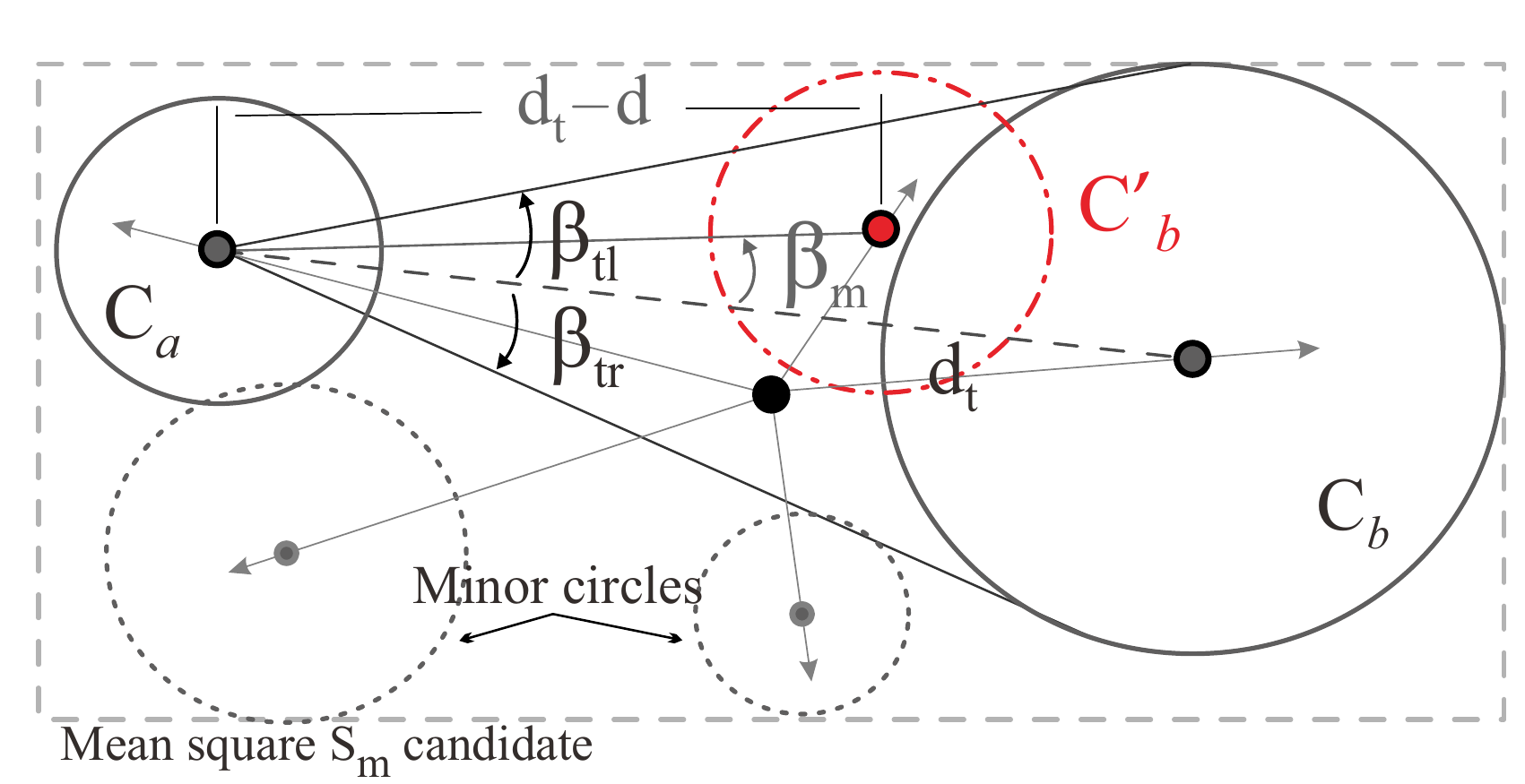}
    \caption{An example case showing how the maximum temporary distance $d_t$ decreases during the collection of a circle couple $\{\uvec{C}_a,\uvec{C}_b\}$.}
    \label{fig:DmFinding}
\end{figure}

A condition has to be constructed to update the temporary distance of $d_{t}$. Also, the algorithm should break a couple of circles and let the expert search for a smaller region to find the right couple of circles as in Figure \ref{fig:DmFinding}. The main reason is the existence of the far objects as $\uvec{C}'_b$ that there is located between/inside the objects (they can be in same distance), here are the matched $\{\uvec{C}_a,\uvec{C}_b\}$. Thus, if the filter combines the two aligned objects, it will create the misinterpretation in object detection and maybe the furthest object would be ignored. Based on the obtained insight, the condition for decreasing the maximum temporary distance $d_{t}$ is determined
\begin{align}
\left[V_{c',b} \leq V_{c,a} \right]\; \& \;
\big[ |\beta_{tr}| \leq |\beta_{tl}| \leq |\beta_{m}| \big]    \; \& \; \left[ d<d_{t} \right]
\label{Eq:ConditionCoupleCirclematchlater}
\end{align}
where $V_{c',b}$ is the candidate circle velocity and 
\begin{equation*}
    \begin{split}
 & \beta_{tl}=\sin^{-1}\left(\frac{||\uvec{L}_{tl}-\uvec{L}_{c,b}||}{||\uvec{L}_{c,a}-\uvec{L}_{c,b}||} \right)  ,\\
&\; \beta_{tr}=\sin^{-1}\left(\frac{||\uvec{L}_{tr}-\uvec{L}_{c,b}||}{||\uvec{L}_{c,a}-\uvec{L}_{c,b}||} \right),\\
&\beta_{m}=\cos^{-1}\left(\frac{d^2_{ab}+d^2_{ab'}-d^2_{bb'}}{2d_{ab}d_{ab'}} \right)
    \end{split}
\end{equation*} 
with
\begin{align*}
    \begin{split}
&d_{ab}=||\uvec{L}_{c,a}-\uvec{L}_{c,b}||,\;d_{a b'}=||\uvec{L}_{c,a}-\uvec{L}_{c',b}||,\;\\
&d_{b b'}=||\uvec{L}_{c,b}-\uvec{L}_{c',b}||
    \end{split}
\end{align*} 
where $\uvec{L}_{c',b}$, $\uvec{L}_{tl}$ and $\uvec{L}_{tr}$ are the locations of the potential circle $\uvec{C}_{b'}$, the left and right tangent points on $\uvec{C}_{b}$ circle.
If the condition (\ref{Eq:ConditionCoupleCirclematchlater}) is satisfied, the temporary maximum distance $d_{t}$ is updated by $d_{t}=d_{t}-d_{bb'}$.

As searching loop continues for matching the circles, the algorithm checks for case 2 matches. These matches are circles that are aligned with approximately similar velocities and direction of motion where they satisfy following condition:
\begin{equation}
\begin{split}
&\left[V_{c,b}-\varepsilon_v\leq V_{c,a}\leq V_{c,b}+\varepsilon_v \right]\\ \; &
\& \;
\big[ \left ( \beta_{c,b}-\delta_{\beta,2}-\varepsilon_{\beta} \right) \leq \beta_{c,a}\leq \left ( \beta_{c,b}+\delta_{\beta,2}+\varepsilon_{\beta} \right) \big] \; 
\end{split}
\end{equation}
where $\delta_{\beta,2}$ is the angle difference for satisfying case two.
 \subsubsection{Mean Square} 
After a successful match for the couple of circles $\{$$\uvec{C}_a$,$\uvec{C}_b$$\}$, the algorithm includes the matching other minor circles into $\uvec{C}_b$ [see Figure \ref{fig:DmFinding} as an example]. The constructed mean square $\uvec{S}_{m}$ from matched circles with included minor circles has following parameters
  \begin{align}
      &\uvec{O}_{m}= \frac{1}{2}\left( \uvec{O}_{c,a}+\frac{1}{l}\sum^l_{i=1} \uvec{O}^i_{c,b} \right),\\
      \label{Eq:squaremakingmeanfirst}
      &\beta_{m}= \angle \left(\uvec{L}_{c',b},\uvec{O}_{s,m} \right),\\
      &V_{m}=  \frac{1}{2}\left(V_{c,a}+\frac{1}{l}\sum^l_{i=1} V^i_{c,b}
      \right),\\
      &\uvec{R}_{m}=(\uvec{L}^+_{m}-\uvec{L}^-_{m})/2,\; \uvec{L}_{m}=(\uvec{L}^+_{m}+\uvec{L}^-_{m})/2
      \label{Eq:squaremakingmean}
  \end{align}
  where
  \begin{align*}
        \uvec{L}^+_{m}= \max \{\uvec{L}_{c,a},\uvec{L}_{c,b} \},~ \uvec{L}^-_{m} = \min\{\uvec{L}_{c,a},\uvec{L}_{c,b}  \}
  \end{align*}
 \makehighlight{Additionally, the algorithm finds the matches of the circles for a minor circle and the circle couple $\{\uvec{C}_{a},\uvec{C}_{b}\}$ is
 \begin{equation} 
\begin{split}
 & \left[ \left ( V_{m}-\varepsilon_v \right)\leq V_{c',b}\leq \left ( V_{m}+\varepsilon_v \right)  \right] \;\& \\ 
  &\;\left[ \left ( \beta_{m} -\varepsilon_{\beta} \right) \leq\beta_{c',b} \leq  \left( \beta_{m}+\varepsilon_{\beta} \right) \right]
         \end{split} \label{Eq:MinorCircleCondition}
 \end{equation}}
\makehighlight{ By succeeding the condition (\ref{Eq:MinorCircleCondition}), if minor circles are located inside the constructed square $ \uvec{L}_{c'}$ $\in$$ \{\uvec{L}^+_{m}$,$ \uvec{L}^-_{m}\}$ by $\{\uvec{C}_a$, $\uvec{C}_b \}$, they are added to the $\uvec{C}_b$. } Note that our $\uvec{C}_a$ is a single candidate but $\uvec{C}_b$ might have more than one circle. This reduces the computation of the search in the algorithm. Also, if maximum temporary distance $d_t$ is required to be reduced in any case, the excluded circles (they are in $d>d_t$ region) are removed from $\uvec{C}_b$. 
  
\subsubsection{Square Estimation}
 \begin{figure}
	\centering
	\includegraphics[width=0.72\columnwidth]{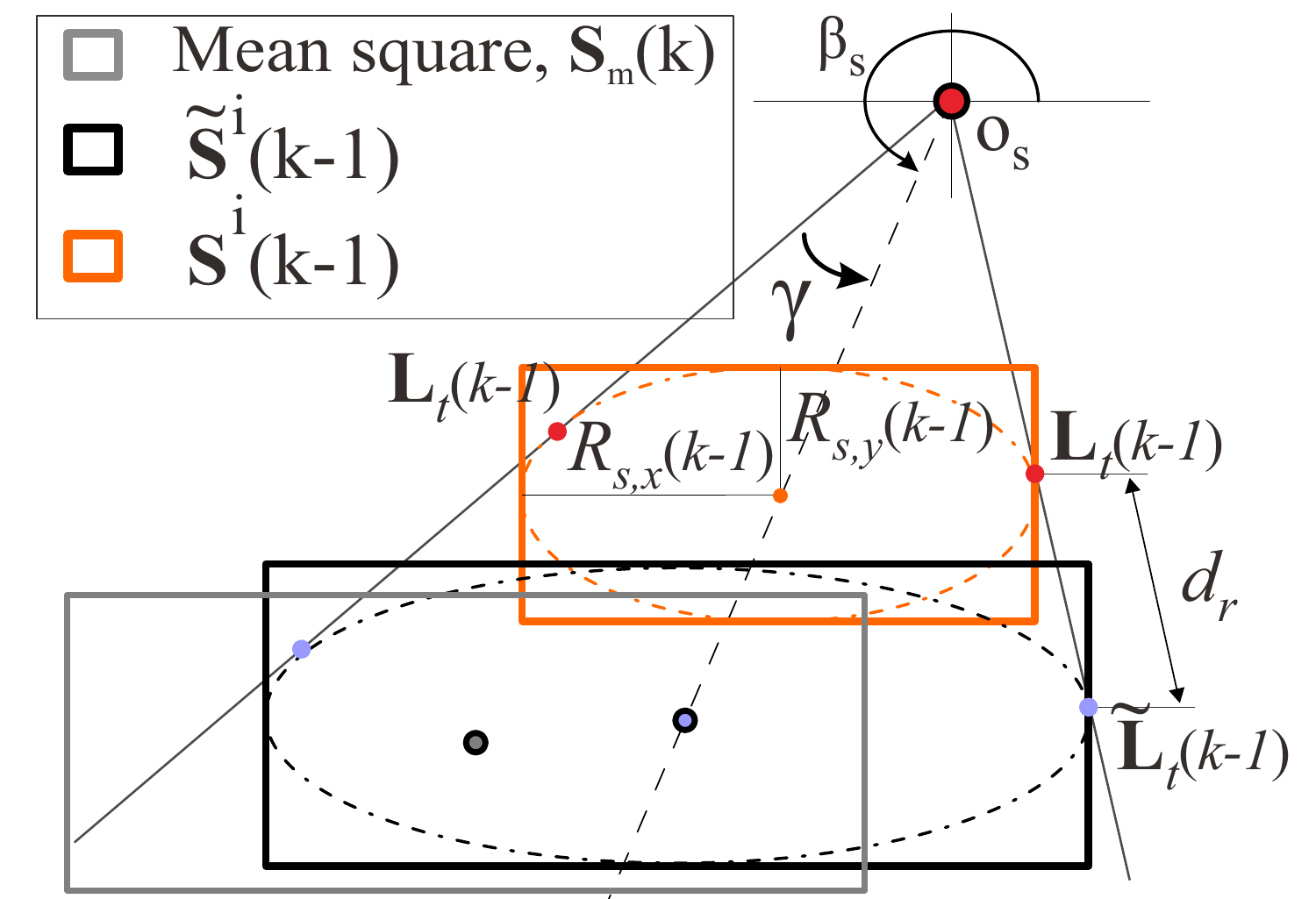}
	\caption{Matching squares with geometric parameters.}
	\label{Fig:Squareestimation}
\end{figure}
After collecting circles that matched with proposed cases, the constructed temporary square $\uvec{S}_m$ by (\ref{Eq:squaremakingmean}) has to be estimated with previous step's prediction that exists at $\uvec{S}$; Hence, we predict $k$-th step's square from $k-1$ step and compare it with constructed temporary one. 

To calculate the predicted squares $\tilde{\uvec{S}}$, the tangent lines are firstly found on the trapped ellipse in each rectangle (called as "square") $\uvec{S}^i(k-1)$ [see Figure \ref{Fig:Squareestimation}] that are derived by solving ellipse and tangent lines' equations for obtaining the couple of points $(L_{t,x},L_{t,y})$,
 \begin{align}
&(L_{t,y}-L_{s,y})^2/R_{s,y}^2+(L_{t,x}-L_{s,x})^2/R_{s,x}^2=1 \label{Eq:Elipticequa}
\\
&R_{s,x}^2(O_{s,y}-L_{t,y})(L_{t,y}-L_{s,y}) \nonumber\\
&+R_{s,y}^2(O_{s,x}-L_{t,x})(L_{t,x}-L_{s,x})=0 
\label{Eq:ElipticTangent}
 \end{align}
Note that these equations have two solutions, so $(L_{t,x},L_{t,y})$ is chosen arbitrarily from the couple for simplicity. Equation (\ref{Eq:ElipticTangent}) is derived by differentiation equation (\ref{Eq:Elipticequa}) as
\begin{align}
dL_y (L_{t,y}-L_{s,y})/R_{s,y}^2+ dL_x(L_{t,x}-L_{s,x})/R_{s,x}^2=0   
\label{Eq:Differentialtangentpointol}
\end{align}
Next, the tangent line slope relation between $(O_{s,x},O_{s,y})$ and $(L_{t,x},L_{t,y})$ points is defined by
\begin{eqnarray}
dL_y/dL_x=(O_{s,y}-L_{t,y})/(L_{t,x}-L_{s,x}),
\label{Eq:Diffpointtangent}
\end{eqnarray}
which substituting (\ref{Eq:Diffpointtangent}) into (\ref{Eq:Differentialtangentpointol}) results in the equation (\ref{Eq:ElipticTangent}). 
 
Before determining the predicted location of the tangent points $\tilde{\uvec{L}}_t(k-1)$ at frame $k$, the traveled distance $d_r$ is found through trigonometric relations as Figure \ref{Fig:Squareestimation}:
 \begin{equation}{\small
d_r= \begin{cases}
   &  ( || \uvec{O}_s-\uvec{L}_s|| +\Delta r)\cos\gamma-|| \uvec{O}_s-\uvec{L}_t||, \; \frac{|| \uvec{O}_s-\uvec{L}_s||}{|| \uvec{O}_s-\uvec{L}_t||}
   > 0\\
  & \frac{|| \uvec{O}_s-\uvec{L}_s||+\Delta r}{\cos \gamma }  - || \uvec{O}_s-\uvec{L}_t||\cos \gamma, \;\;\; \;\;\;\;\;\;\;\;\;\hfill \frac{|| \uvec{O}_s-\uvec{L}_s||}{|| \uvec{O}_s-\uvec{L}_t||}
   \le 0
 \end{cases}}
 \label{Eq:distancetangent}
 \end{equation}
where $\Delta r=||\uvec{L}_{s}(k-1)-\tilde{\uvec{L}}_{s}(k-1)||$ is the distance difference between the location of square in the frame $k-1$ and the predicted location in $k$-th step that is calculated by solving the vehicle kinematics, and $\gamma$ is
 \begin{equation}
  \gamma=\cos^{-1}
  \begin{cases}
 \begin{split}
&\frac{|| \uvec{O}_s-\uvec{L}_t||}{|| \uvec{O}_s-\uvec{L}_s||},\;\;\; || \uvec{O}_s-\uvec{L}_s||>|| \uvec{O}_s-\uvec{L}_t||\\
&\frac{|| \uvec{O}_s-\uvec{L}_s||}{|| \uvec{O}_s-\uvec{L}_t||} ,\;\;\;||\uvec{O}_s-\uvec{L}_s|| \leq || \uvec{O}_s-\uvec{L}_t||
 \end{split}
 \end{cases}
 \label{Eq:distanceangletangent}
 \end{equation}
 Note that all the parameters in these formulas are related to $k-1$ frame such as $\uvec{L}_s=\uvec{L}_s(k-1)$ and $\uvec{L}_t=\uvec{L}_t(k-1)$. We have wrote them this way for simplicity. By substituting equation (\ref{Eq:distanceangletangent}) to equation (\ref{Eq:distancetangent}), the traveled distance $d_r$ is simplified to
  \begin{equation} {\small
  d_r=
 \begin{cases}
   &  ( || \uvec{O}_s-\uvec{L}_s|| +\Delta r)\frac{|| \uvec{O}_s-\uvec{L}_t||}{|| \uvec{O}_s-\uvec{L}_s||}-|| \uvec{O}_s-\uvec{L}_t||, \hfill \frac{|| \uvec{O}_s-\uvec{L}_s||}{|| \uvec{O}_s-\uvec{L}_t||}
   > 0 \\
  & (|| \uvec{O}_s-\uvec{L}_s|| +\Delta r)\frac{|| \uvec{O}_s-\uvec{L}_s||}{|| \uvec{O}_s-\uvec{L}_t||} - || \uvec{O}_s-\uvec{L}_s||,  \hfill \frac{|| \uvec{O}_s-\uvec{L}_s||}{|| \uvec{O}_s-\uvec{L}_t||}
   \le 0
 \end{cases}
 \label{Eq:distancetangent2}}
 \end{equation}
\noindent Next, the slope for these couple of points are determined by 
\begin{align}
    m_{t}=(L_{t,x}(k-1)-O_{s,x})/(L_{t,y}(k-1)-O_{s,y})
\end{align}
Now, the predicted location of the tangent couple of points $\tilde{\uvec{L}}_t$ at the frame $k$ are
\begin{align}
    \begin{split}
        &\tilde{L}_{t,y}(k-1)= L_{t,y}(k-1)+\left(\frac{d^2_r}{m_t^2+1}\right)^{\frac{1}{2}}, \\
        &\tilde{L}_{t,x}(k-1)= L_{t,x}(k-1)+m_t[L_{t,y}(k)- \tilde{L}_{t,y}(k-1)]
    \end{split}
    \label{Eq:Findingthekthsteppoints}
\end{align}
The same equation (\ref{Eq:Findingthekthsteppoints}) is used to predict the location of the considered square candidate $\uvec{L}_s(k)$ in $k$-th frame. Finally, the new minor and major radii of the ellipse as the sides of the geometric rectangle (the square) are found by solving following algebraic equations: 
\begin{equation}{\footnotesize
 \begin{split}
     &\tilde{R}_{s,y}(k-1)= \Big| (O_{s,x}-\tilde{L}_{t,x}(k-1))(\tilde{L}_{t,y}(k-1)-\tilde{L}_{s,y}(k-1))^2 \\
     & - (O_{s,y}-\tilde{L}_{t,y}(k-1))(\tilde{L}_{t,x}(k-1)-\tilde{L}_{s,x}(k-1))(\tilde{L}_{t,y}(k-1)\\
     &-\tilde{L}_{s,y}(k-1))/\left[O_{s,x}-\tilde{L}_{t,x}(k-1) \right]\Big|^{\frac{1}{2}} \\
      &\tilde{R}_{s,x}(k-1)=\Big| \tilde{R}^2_{s,y}(k-1)(O_{s,x}-\tilde{L}_{t,x}(k-1))(\tilde{L}_{t,x}(k-1)\\
      &-\tilde{L}_{s,x}(k-1))/\left[(O_{s,y}-\tilde{L}_{t,y}(k-1))(\tilde{L}_{t,y}(k-1)-\tilde{L}_{s,y}(k-1)) \right] \Big|^{\frac{1}{2}} 
\end{split}}
\label{Eq:minormajorpredictedsquare}
\end{equation}

The velocity of the predicted $i$-th square $\tilde{\uvec{S}}^i(k-1)$ with the known $\tilde{\uvec{L}}^i_s(k-1)$ and $\tilde{\uvec{R}}^i_s(k-1)$ is $\tilde{V}^i_s(k-1)=V^i_s(k-1)$. Note that velocities of the Circle and Square experts from previous frame $(k-1)$ are always updated with new vehicle velocity $V_v$ if the mobile robot is not moving in constant speed. Now, the predicted square $\tilde{\uvec{S}}^i(k-1)$ has to be compared with constructed square $\uvec{S}_m(k)$ from the collected circles 
\begin{equation}
    \begin{split}
        & [\tilde{V}^i_{s}(k-1)-\varepsilon_{v,s} \leq V_{s,m}(k) \leq \tilde{V}^i_{s}(k-1)+\varepsilon_{v,s} \;  ] \\
        & \& \; [\tilde{\beta}_s(k-1)-\varepsilon^i_{\beta,s} \leq \beta_{s,m}(k) \leq \tilde{\beta}^i_s(k-1)-\varepsilon_{\beta,s} \; ]\\
      & \& \; [ \rho_{s}> \rho_{c}]
    \end{split}
    \label{Eq:conditionofmatchingtwosquare}
\end{equation}
\noindent where $\rho_{s}$, $\rho_c$, $\varepsilon_{\beta,s}$ and $\varepsilon_{v,s}$ are the overlap percentage between predicted and constructed squares, minimum overlap constant, the accuracies of the angle and velocity, respectively. The overlap percentage $\rho_{s}$ between the predicted and constructed squares is
\begin{equation}
     \rho_{s}=\frac{100\cdot\;S_{oa}}{4\cdot \min\{\tilde{R}_{s,x}(k-1)\cdot\tilde{R}_{s,y}(k-1),R_{m,x}(k)\cdot R_{m,y}(k)\}} 
\end{equation}
where the overlap area $S_{oa}$ is the constructed rectangle geometry between predicted rectangle $\tilde{\uvec{S}}(k-1)$ and the constructed rectangle by circles in $\uvec{S}_m$ is defined by 
\begin{align}
    S_{oa}=L_{oa,x}\cdot L_{oa,y}
\end{align}
where
\begin{equation*}
\begin{split}
& L_{oa,x}=\big |\min\{\tilde{L}_{s,x}(k-1)+\tilde{R}_{s,x}(k-1),L_{m,x}(k)+R_{m,x}(k)\}\\
&-\max\{\tilde{L}_{s,x}(k-1)-\tilde{R}_{s,x}(k-1),L_{m,x}(k)-R_{m,x}(k)\}\big|,\\
& L_{oa,y}=\big |\max\{\tilde{L}_{s,y}(k-1)-\tilde{R}_{s,y}(k-1),L_{m,y}(k)-R_{m,y}(k)\}\\
&-\min\{\tilde{L}_{s,y}(k-1)+\tilde{R}_{s,y}(k-1),L_{m,y}(k)+R_{m,y}(k)\}\big |
\end{split}
\end{equation*}
After satisfying the condition (\ref{Eq:conditionofmatchingtwosquare}), the parameters of the square filter are estimated by using equation~(\ref{Eq:estimationgeneralfor}). Please note that for the initialization all squares are constructed by the collected mean squares in equations~(\ref{Eq:squaremakingmeanfirst})-(\ref{Eq:squaremakingmean}). Also, the trust factor is updated similarly to the normal circle definitions. 
\begin{figure*}
	\centering
 	\includegraphics[width=  3.6 in, height=3.0 in]{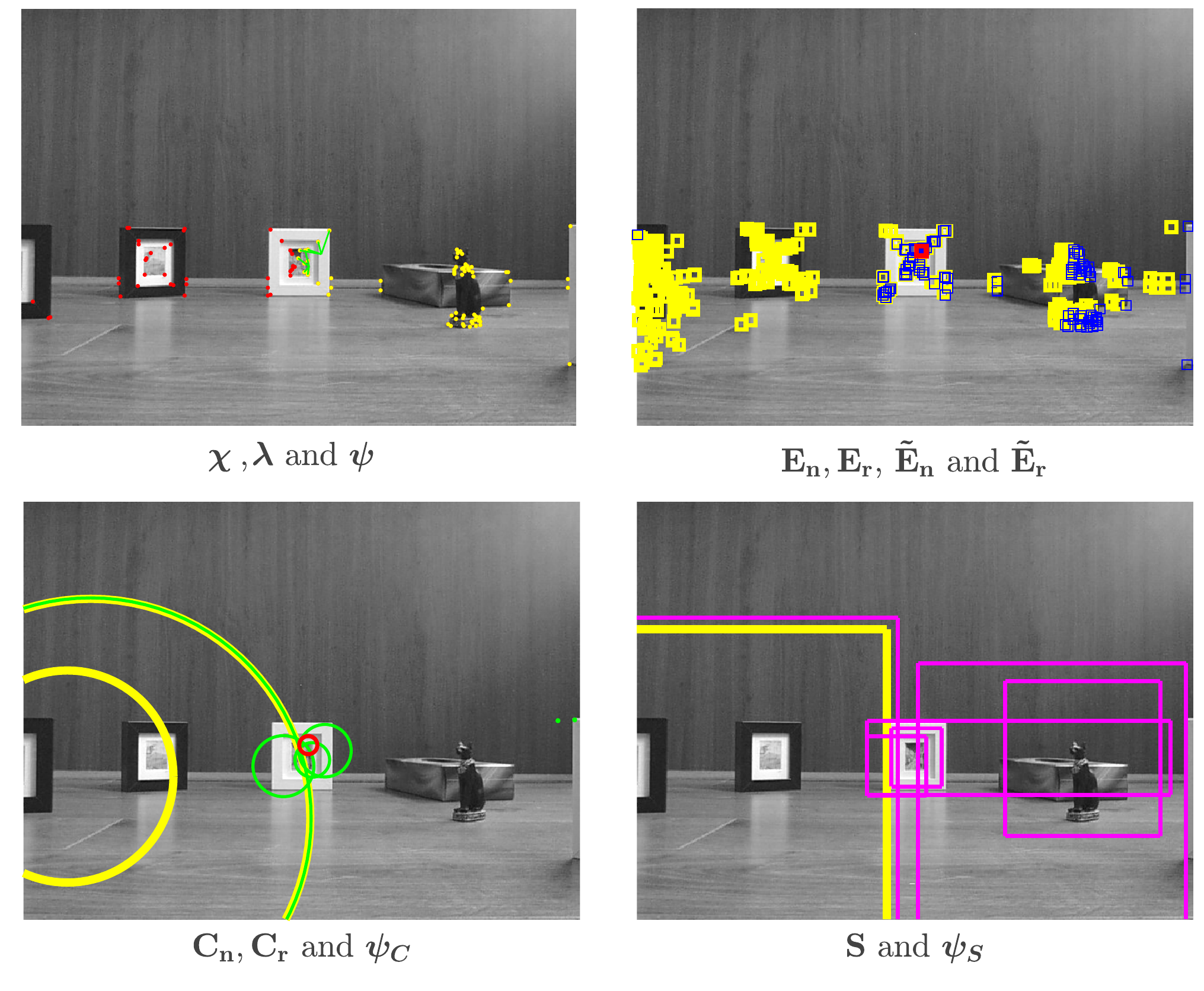}
	\caption{Example frame ($k=9$) with filter experts. Top left: red and yellow dots are showing detected $\pmb{\chi}$ and ignored edges $\pmb{\psi}$, line presents the line spacing $\pmb{\lambda}$; Top right: the yellow and blue squares are approximated edge $\tilde{\uvec{E}}_n,\tilde{\uvec{E}}_r$ and estimated normal edge $\uvec{E}_n$ and red square is rebel edges $\uvec{E}_r$; Bottom left: green, red and yellow colors are for normal circles $\uvec{C}_n$, rebel circles $\uvec{C}_r$ and ignored circles $\pmb{\psi}_C$; Bottom right: magenta and yellow rectangle are the squares $\uvec{S}$ and ignored ones $\pmb{\psi}_S$. }
	\label{Fig:sIZEDImention}
\end{figure*}

%% file: contents/evaluation.tex
\section{Evaluation} \label{sec:evaluation}

 
\makehighlight{In order to evaluate the performance and the properties of the proposed LCS filter, we first explain the filter in a simple scenario in Section 5.1, and then we check how individual expert works. Also, memory overhead and computational efficiency are evaluated. Next, we apply the filter for autonomous vehicles in Section 5.2. Finally, the filter's performance is checked with real-world scenarios in autonomous cars on a sunny day and a rainy day at night with a high threshold for corner detection.}

\subsection{\makehighlight{Filter Behavior in Simple Scenario}}
The experiment is set up as follows: an IMU sensor and a high-definition camera (with a resolution of $640 \times 480$ pixels) are attached to a mobile robot platform to collect image and motion data which are fed to the LCS algorithm.
Our study takes place with the worst-case scenario where the frame rate is at 1 frame/s. Although our geometric filter demonstrates its best performance with at least 3 frame/s, due to low velocity ($V_v$ = $3$ cm/s) and minor acceleration $a_v=0.05$ cm/s$^2$ of the robot. A corner detector, \textit{FAST9} \cite{Rosten2006,Rosten2010}, is used as the detection back-end, with 20 point threshold utilized. All the filter parameters are given in Table \ref{Tab:LCSFilterparameters}. It is assumed that there are minor deviations ($\delta_v=9^o$) in the robot motion. We have chosen some of the values for the parameters (for example the accuracy parameter $\varepsilon$) with respect to the environment geometric complexity, object detection accuracy, and processing speed. For instance, if the environment is very complex, the angular and velocity accuracy of the Circle ($\varepsilon_{\beta,n},\varepsilon_{\beta,r},\varepsilon_{v,n},\varepsilon_{v,r}$) and the Square experts ($\varepsilon_{\beta,1},\varepsilon_{\beta,2},\varepsilon_{\beta,s},\varepsilon_{v,s}$) variables should be re-tuned (certain increase) to prevent over-computation. By increasing accuracy (lower $\varepsilon$), we force the filter to separate landmarks with a higher possible degree which means more memory usage. Also, the trust factors are chosen in given values to let the square candidates stay longer than circles due to their abstract and high-level object detection.

The environment is chosen crowded with an average of 75-200 landmarks in each frame. The linear kinematics of the vehicle is solved using MATLAB \textit{ode45} solver to predict the location of landmarks and experts (circle and square) as well as the $\Delta r$ in equation (\ref{Eq:distancetangent}). Please note that the superscripts $c$ and $s$ in trust factors $Tr$ in Table \ref{Tab:LCSFilterparameters} stand for the Circle (edges and circles) and Square experts. 
 
\begin{table*}
	\renewcommand{\arraystretch}{1}
	\caption{Parameters for the LCS filter used in the experiment.}
	\label{Tab:LCSFilterparameters}
	\centering
		\begin{tabular}{cc|cc|cc}
			\hline
			\hline
			Variable & Value & Variable & Value &  Variable & Value\\
			\hline
			$O_{I,x}$ & $320$ &$Tr^c_s$ & 3 & $\varepsilon_{\beta,1}$ & 15 \\
			$O_{I,y}$ & $240$ & $Tr^c_c$  &  2& $\varepsilon_{\beta,2}$ & 35\\
			$V_v$&$3$ cm/s & $Tr^c_m$ & 5 & $\varepsilon_{\beta,s}$ & 20 \\
            $a_v$&$0.05$ cm/s$^2$ 	& $Tr^s_s$ & 5& $\varepsilon_{\beta}$  & 20\\
			$\delta_v$&$9^o$ &$Tr^s_c$ & 3& $\varepsilon_{v,r}$ & 100\\
			$\delta_{\beta}$& $90^o$& $Tr^s_m$ & 7& $\varepsilon_{v,s}$& 0.7\\
			$\mu_0$& $25$ & $\varepsilon_{\beta,n}$  & 20 &$\varepsilon_{v,n}$ & 40 \\
			$\rho_c$&  40& $\varepsilon_{\beta,r}$ & 50 &  $\varepsilon_{v}$ &  0.7\\
			\hline
			\hline
	\end{tabular} 
\end{table*}
At first, we show an example frame as in Figure \ref{Fig:sIZEDImention} to explain how the results are interpreted in each expert and its geometric presentation. 

 \begin{figure*}
	\centering
	\includegraphics[width=4.6 in, height=1.3 in ]{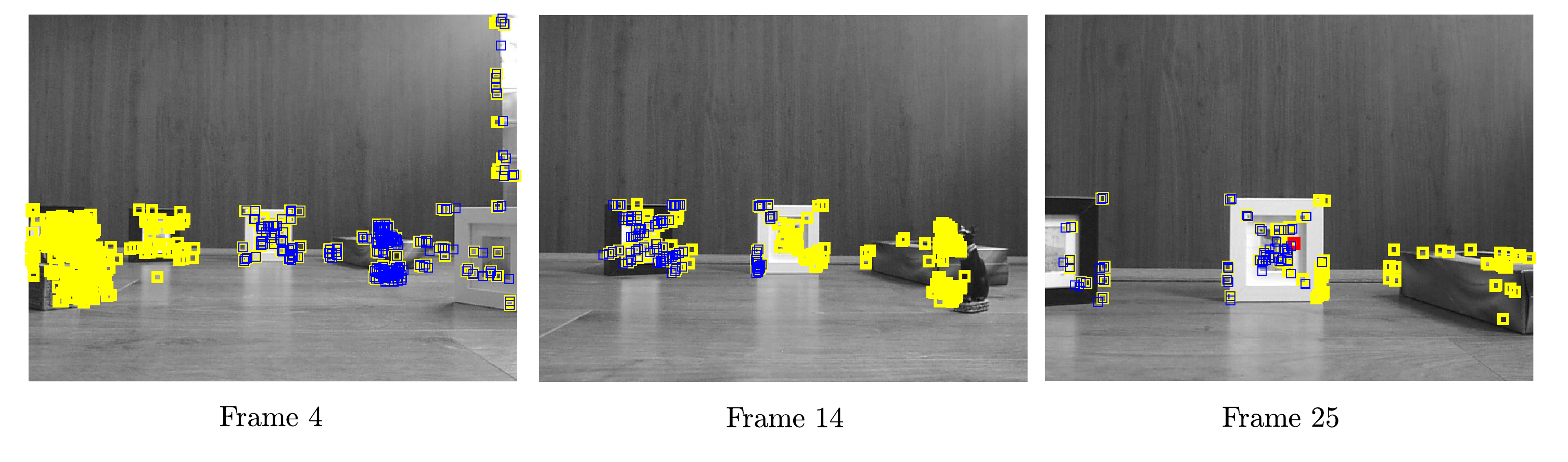}
	\includegraphics[width=4.6 in, height=1.3 in ]{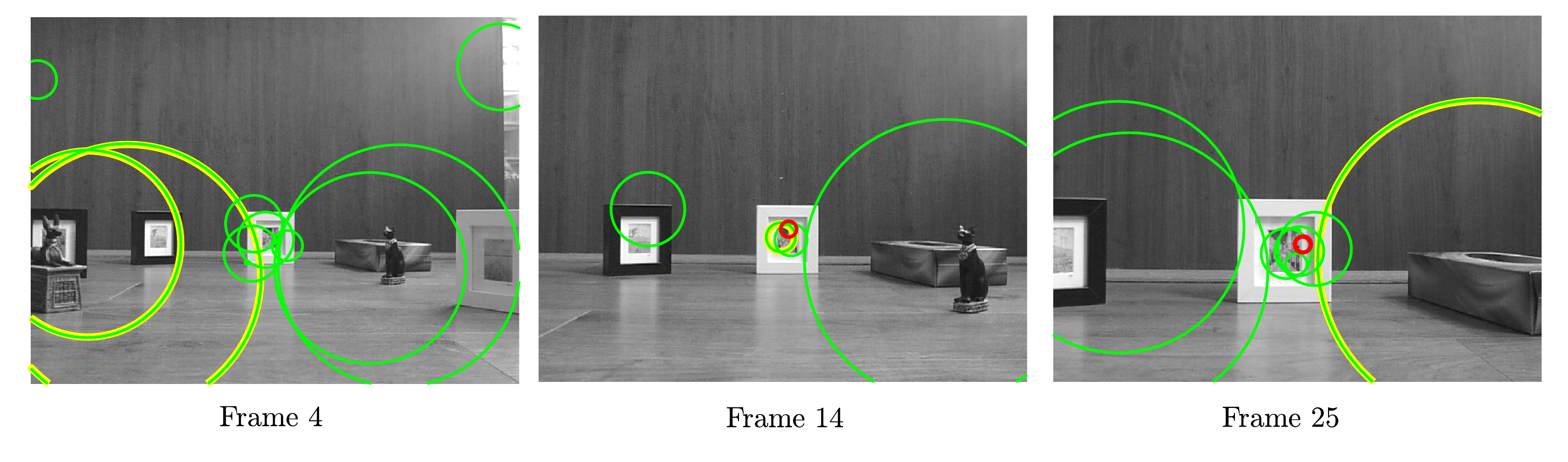}
	\includegraphics[width=4.6 in, height=1.3 in ]{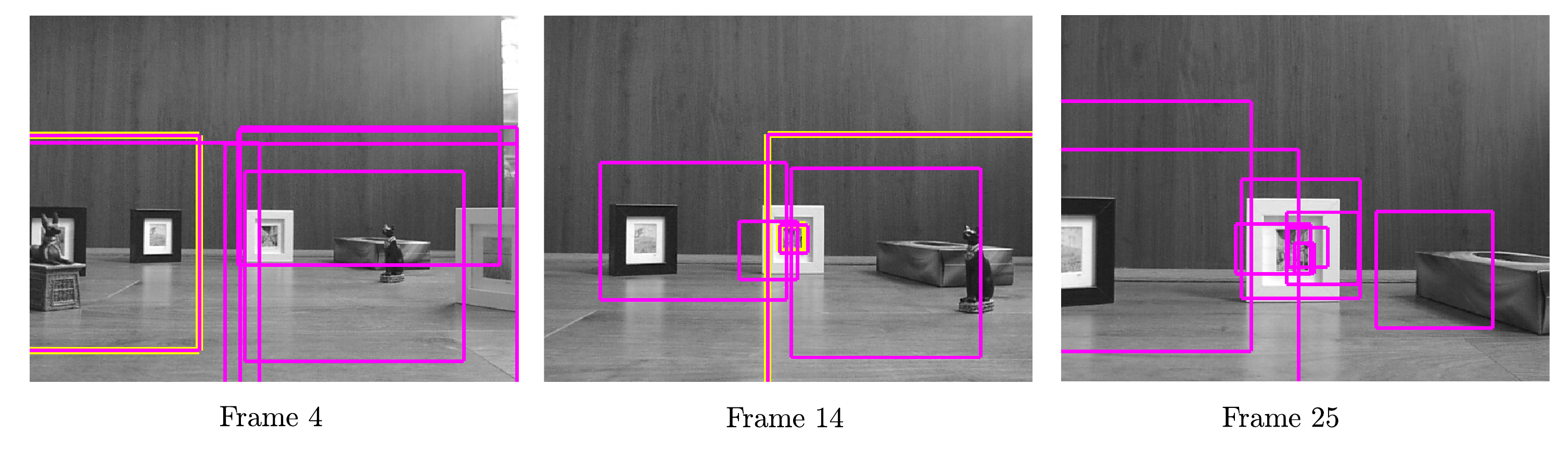}
	\caption{Comparison between sampled frames for understanding the Circle and Square experts behavior.}
	\label{Fig:Compareframes}
\end{figure*}

We have collected a total of 37 frames during the motion of the robot. The full video is available in our open-source code. We have located certain objects (statue) with different forms as well as frame objects that contain complex graphical presentations as shown in Figure \ref{Fig:sIZEDImention}. Each frame consists of four identical images and each shows specific parameters of the LCS filter. At first, the top-left hand-side frame presents the operation of the Line expert [see Figure \ref{Fig:LineExpert} for the flowchart of the Line expert] with detected landmarks $\pmb{\chi}$ in yellow dots where they were obtained by \textit{FAST9} edge detection algorithm. Red dots are the ones that our filter ignores by $\pmb{\psi}$ which is updated through the Circle and Square experts. Also, $\pmb\lambda$ is the grouping parameter that Line expert does to decrease the computation and find the detection radius of the edges $\mu$ when $\pmb{\chi}$ is transferred to the Circle expert. Then, top-right and bottom-left figures in Figure \ref{Fig:sIZEDImention} is for the Circle expert, the Circle expert computation flow is in Figure \ref{Fig:EdgeEstimation2} and Figure \ref{Fig:CircleEstimation}. The blue and red small squares are the final estimated location of the normal $\uvec{E}_n$ and rebel $\uvec{E}_r$ edges by the expert where they are calculated by equation~(\ref{Eq:estimationgeneralfor}). Note that these red landmarks do not follow the normal vector field flow which means the object either coming toward the camera or is an independently moving object in the frame. The yellow small squares are presenting the predicted squares ($\tilde{\uvec{E}}_n,\tilde{\uvec{E}}_r$) of frame $k-1$ before doing estimation. Next, the circling operation takes place where green and red circles are presenting the normal $\uvec{C}_n$ and rebel $\uvec{C}_r$ circles [see bottom-left image]. These geometries are tracking the layers that have similar relative velocity, location, and angle. The yellow circle stands for the layer that reaches maximum trust $Tr^c_m$ in the frame $k-1$ where the ignorance (Circle expert ignorance update $\pmb\psi_C$) is applied by the update in $\pmb\psi$ at current frame ($k=9$). Finally, the bottom-right figure shows the Square expert operation based on the flowchart in Figure \ref{fig:my_label_SFilter}. Similar to the Circle expert, magenta and yellow rectangle geometries are presenting the estimated squares $\uvec{S}$ and ignored $\pmb\psi_S$ (which get updated from the previous frame) ones. Note that the Square expert combines the layers of the circles to determine more compact objects based on their matching kinematics geometrically. Thus, after every frame passes, the accuracy of estimation gets higher and information becomes more compact and summarized (high-level features) [see Figure \ref{Fig:AlgorithmFunctioningMAP} for the overall filter]. This helps to create a continuous feed to minimize the amount of collected data, e.g., edges.

We check the behavior of the LCS filter with some collected frames ($k=4,\;14$ and $25$) as shown in Figure \ref{Fig:Compareframes}. It is clear that the location of the landmarks in the first row of images is estimated with accuracy where they are shown as small blue squares. However, because we have chosen our study as 1 frame/s in the worst-case scenario, some motion errors create a deviation in predicted landmarks (yellow squares). This can be minimized by having more accuracy in the velocity/angular displacement coming from IMU and increasing the number of frames that are sampled. As a strength of the LCS filter, it is able to detect the rebel edges that stand for incoming objects in the scene. The next row of images is for circles with expressing layers with similar kinematics. As the robot moves closer to the objects, they get distinguished better from each other [see Frame 14 and 25] rather than being in a compact form [see Frame 4]. 
 Next, the Square expert can locate the objects especially the one that is in the middle. Please note that one can increase the trust factor of the Square expert to keep the coming data. We think increasing the trust factors ($Tr^s_s$ and $Tr^s_m$) and decreasing the critical trust ($Tr^s_c$) are beneficial for a higher accuracy. However, in that case, the sensory data (IMU) and the robot kinematics model should be more accurate to prevent over-confidence in the filter. It is important to note that when a certain number of objects are too far away, their velocity/location are unified. Thus, this can be an interesting future work to determine the distance sensitivity of this proposed geometric filter. Also, the algorithm determines the objects with $\delta_{\beta}=90^o$ angular matches here in an arbitrary direction. We plan to see, as future work, how we can transform the square data to detection for more smooth curves on the complex objects, more in their natural curvature (such as the status in the scene). However, the designed filter can be very beneficial for applications including mapping (Monocular SLAM) \cite{mur2017orb} and motion control \cite{wisth2019preintegrated,wisth2019robust} that utilize the image features (edge detection) in their planning model or control law. It can be used as a data reduction filter for the scene to speed up the computation while it is increasing the accuracy of detected landmarks. Also, with specifying properties, e.g. velocity, of certain objects in the frame, we can use the filter to apply accurate object recognition methods \cite{williams2009comparison,galvez2012bags,rublee2011orb} in only certain regions of the frame for recognizing what they are. 
 \begin{figure}
	\centering
	\includegraphics[width=2.8 in]{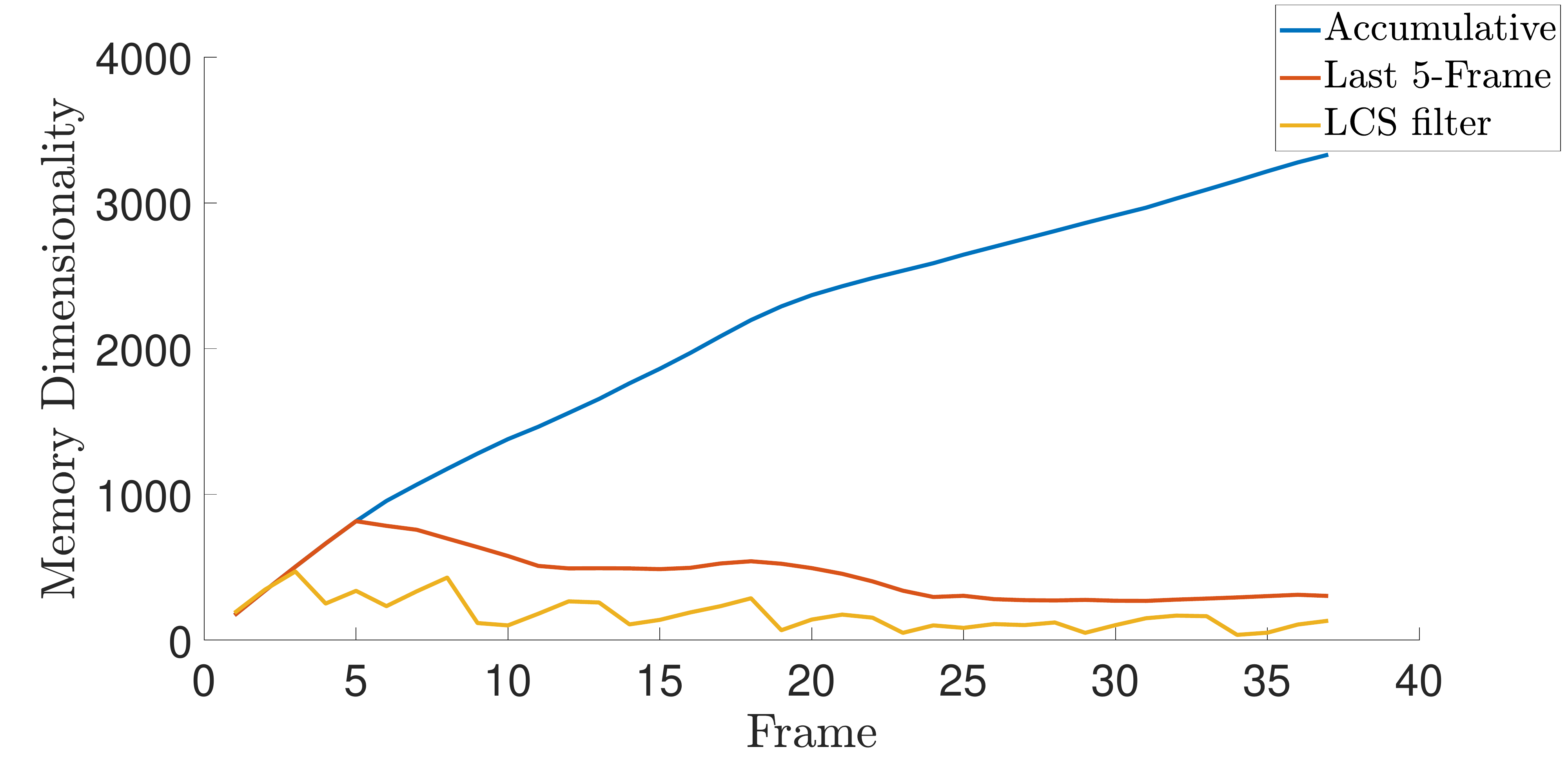}
	\caption{Result of the memory dimensionality verses the frame index.}
	\label{Fig:DimenMem1}
\end{figure}
 \begin{figure}
	\centering
    a)\includegraphics[width=2.8 in]{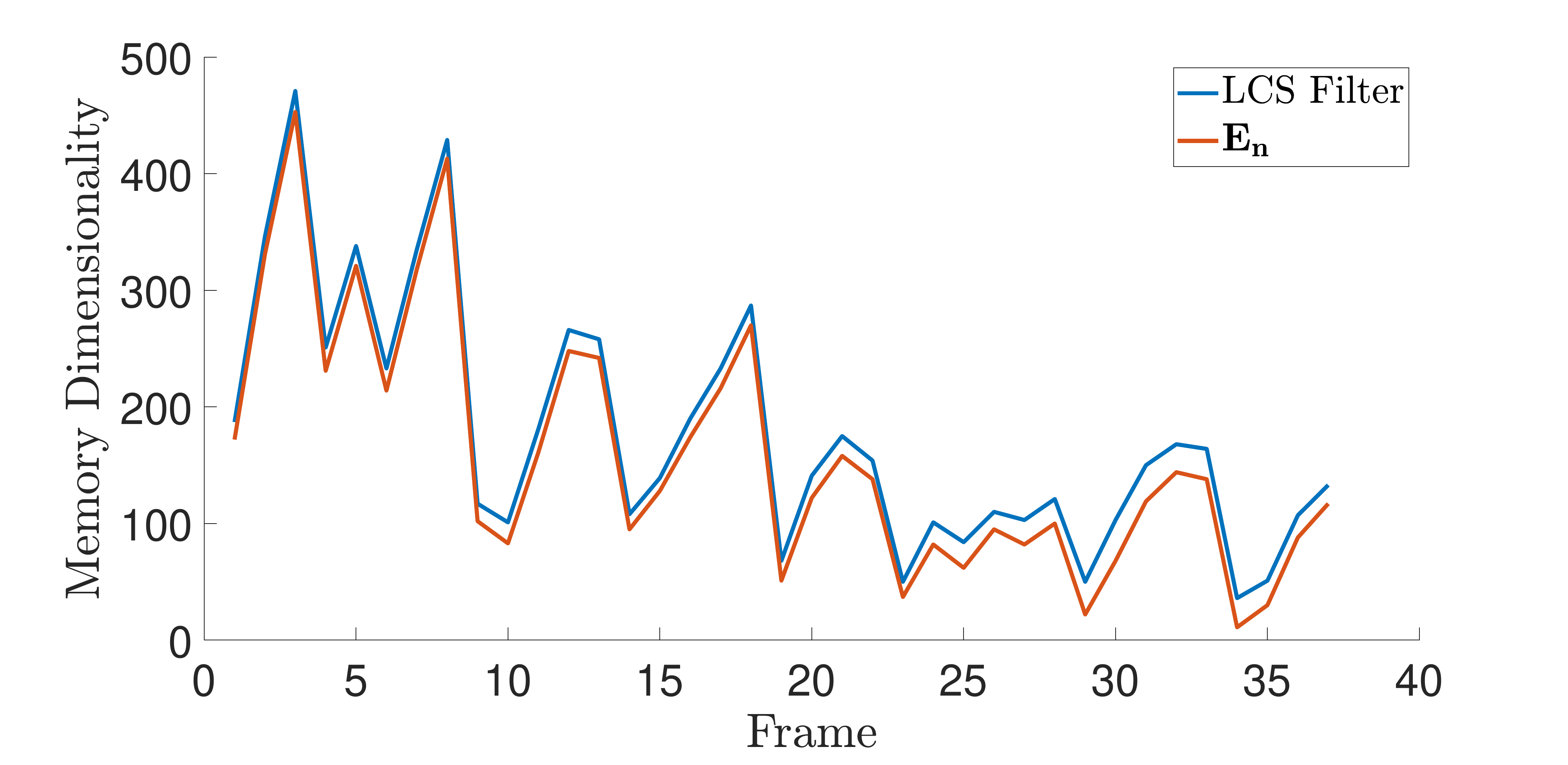}
	b)\includegraphics[width=2.8 in]{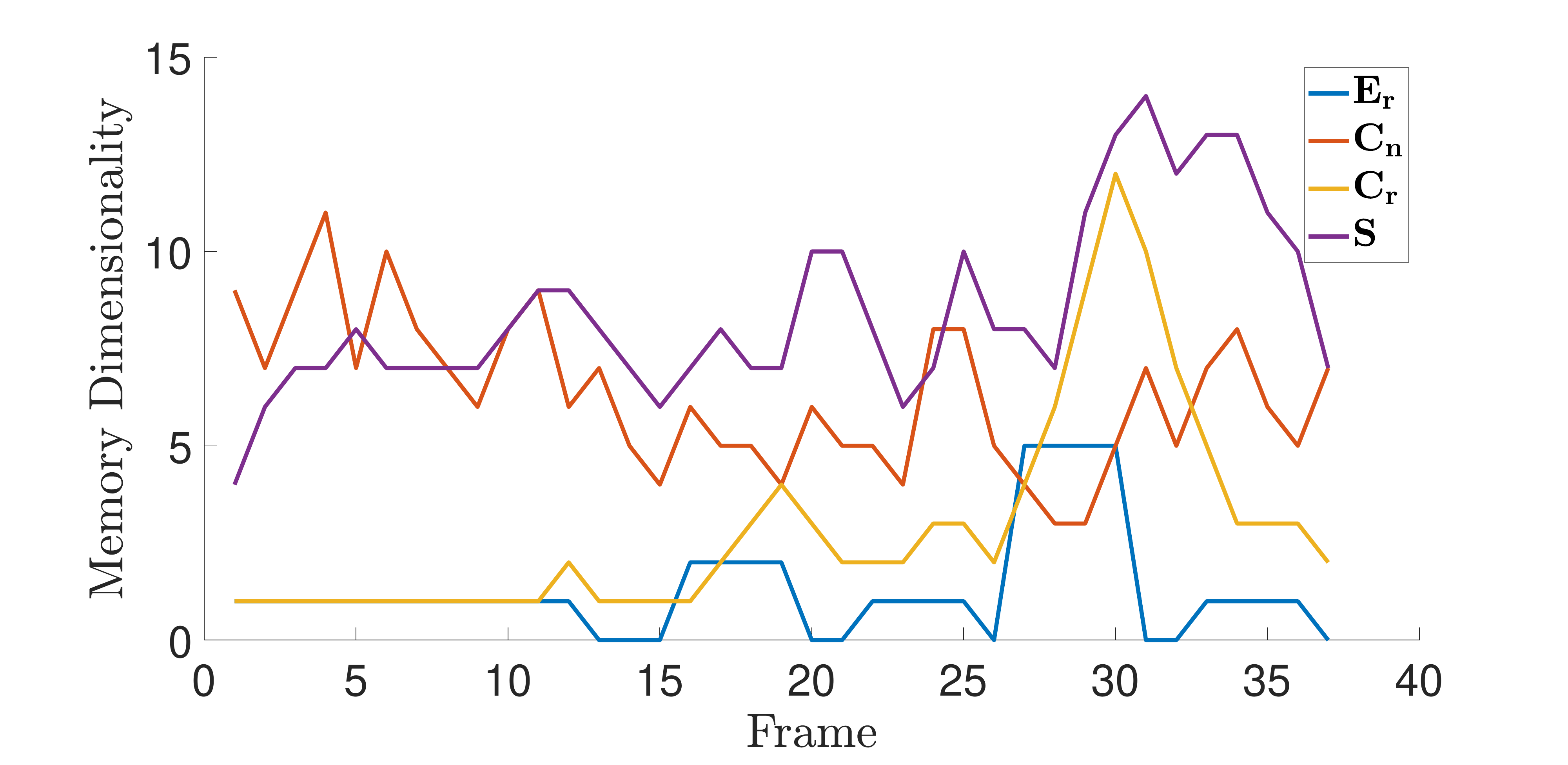}
	\caption{Detailed dimensionality analysis of the LCS filter parameters: a) The occupied size of normal edges $\uvec{E}_n$ in comparison to overall LCS filter, b) The memory occupation for the parameters of the experts in the LCS filter.}
	\label{Fig:DimenMem2}
\end{figure}

Next, we check the performance of our filter with respect to the memory dimensionality and computation complexity. In here, the memory dimensionality means the matrix size. The dimensionality can be looked at as the number of candidates (landmark or layer) each parameter has in the filter (the parameters are in Figure \ref{Fig:AlgorithmFunctioningMAP}) every frame. Figure \ref{Fig:DimenMem1} shows that the LCS filter memory usage based on the dimensionality. We compared our filter with two cases of memory saving: An accumulative data collection that normally takes place in mapping problem and a 5-frame constraint data collection. We can observe that the total size of our filter (including all the expert parameters) is drastically smaller than both of the other cases. Even for the case that the last five frames are saved in memory, our filter still has a half size of its counterpart. Interestingly, as the LCS filter obtains more data in each frame, the dimensionality of it decreases to around 100. In order to have a more detailed analysis of the data flow of each expert in the filter, Figure \ref{Fig:DimenMem2} is presented. The most of data logically are restored in the normal edge $\uvec{E}_n$ matrix since most of the objects are passing with the following vector field of the vehicle [see Figure \ref{Fig:VectorField}-a] motion as it moves with the high number of detected edges. Consequently, the rebel edges $\uvec{E}_r$ and  $\uvec{C}_r$ circles occupy the least memory. However, there is a considerable increase in the number of detected rebel circles $\uvec{C}_r$ as the camera reaches closer to the incoming frame object at the end (when $k=[25-35]$). Additionally, the normal circles $\uvec{C}_n$ and squares $\uvec{S}$ have more dynamic fluctuation in their memory size but the overall trend is approximately constant.

In addition to the reduced memory footprint, the computational complexity of the LCS filter is relatively low. The time complexity of the circle expert (see the flowchart in Figure~\ref{Fig:CircleEstimation}) is correlated to the number of input edges, $n$. The complexity of the circle expert is $\mathcal{O}(2n \cdot log(n))$ as two main search loops used in matching are involved. 
A similar result applies to the Square Expert (see Figure~\ref{fig:my_label_SFilter}), in which the time complexity is based on the number of normal and rebel circles in total.
As the filter is designed in three different layers, it is possible to run the filter in a pipeline fashion which allows for a higher degree of parallelism, e.g., by utilizing multi-cores or general-purpose graphic processing unit (GPGPU). 

\subsection{\makehighlight{Experiment Studies on the Autonomous Cars}}
\begin{figure}
	\centering
 	\includegraphics[width= 3.3 in]{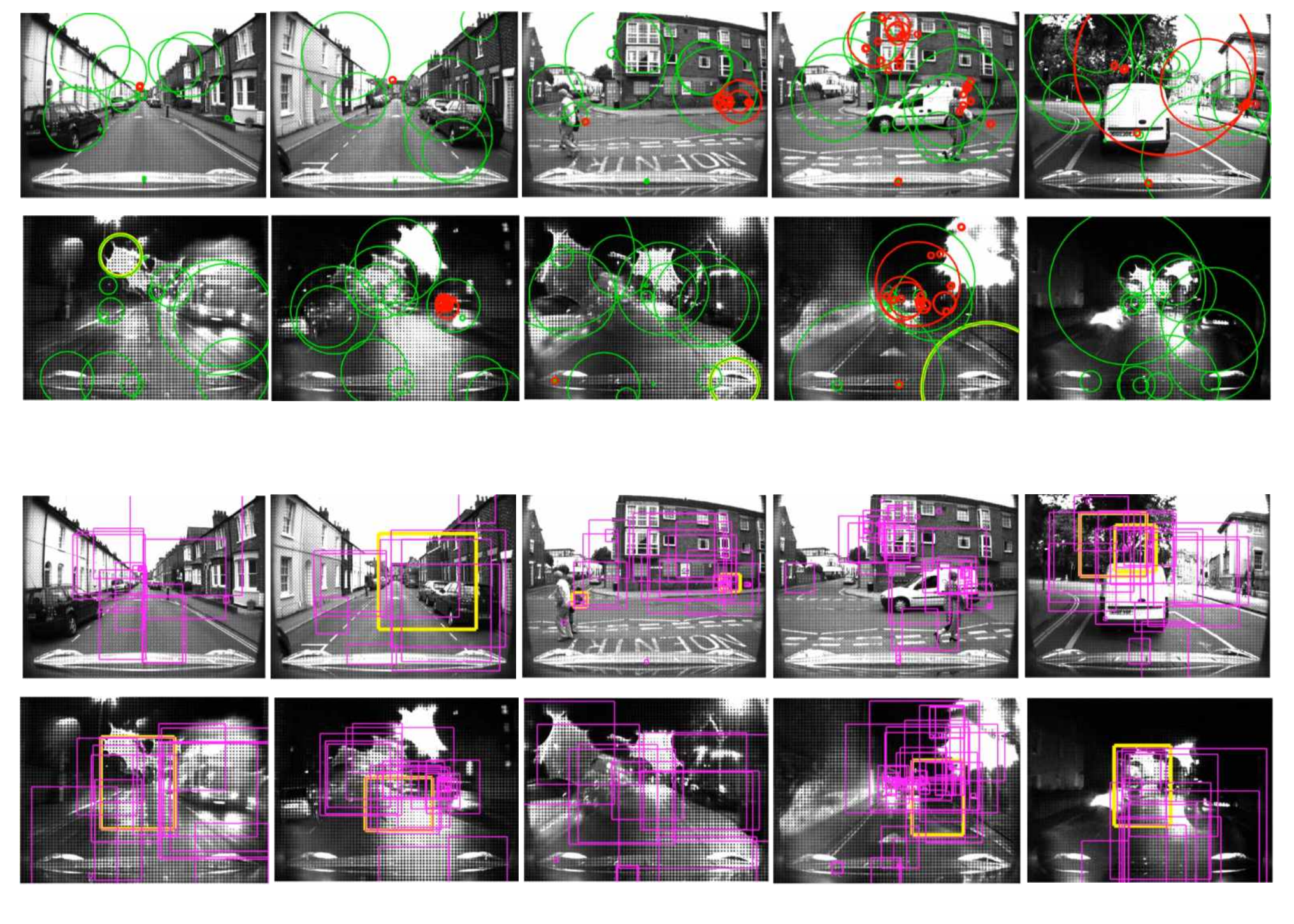}
	\caption{\makehighlight{Example captured frames (day/night) used in the autonomous car example.}}
	\label{Fig:ImageSeries}
\end{figure}
\makehighlight{In this part, we applied our filter in a real-world scenario with autonomous driving cars. The Oxford RoboCar dataset \cite{RobotCarDatasetIJRR} is chosen for this study.}

\makehighlight{Within the dataset, we have chosen scenes that are much more crowded and complex with over 700 detected edges in each frame. The algorithm trust factors are re-tuned as $Tr^c_c=1$, $Tr^c_s=4$ and $Tr^c_m=7$ for the Circle expert and for the Square expert $Tr^c_c=2$, $Tr^s_s=5$ and $Tr^s_m=8$ to have better performance. Other accuracy variables have minor changes where reader can check the open-source code for the details. We have experimented with the following two cases: 1) a scene during day light and 2) a rainy scene at night. Both cases are experimented with over 2200 frames. Figure~\ref{Fig:ImageSeries} depicts some sampled frames from our study. The Circle expert in our filter is able to detect the independently moving objects (rebel landmarks) as pedestrians and passing-by cars as well as normal landmarks. This is true even when it is night with many distorted images due to rain. For example, the filter detects a parking car that moves toward the camera and then parks, see the last image before the end in Figure~\ref{Fig:ImageSeries}. Also, the Square expert can group the layers of the objects in both cases. The long-version videos are included in our source code repository for further check.}

\begin{figure}
	\centering
 	\includegraphics[width= 3.3 in]{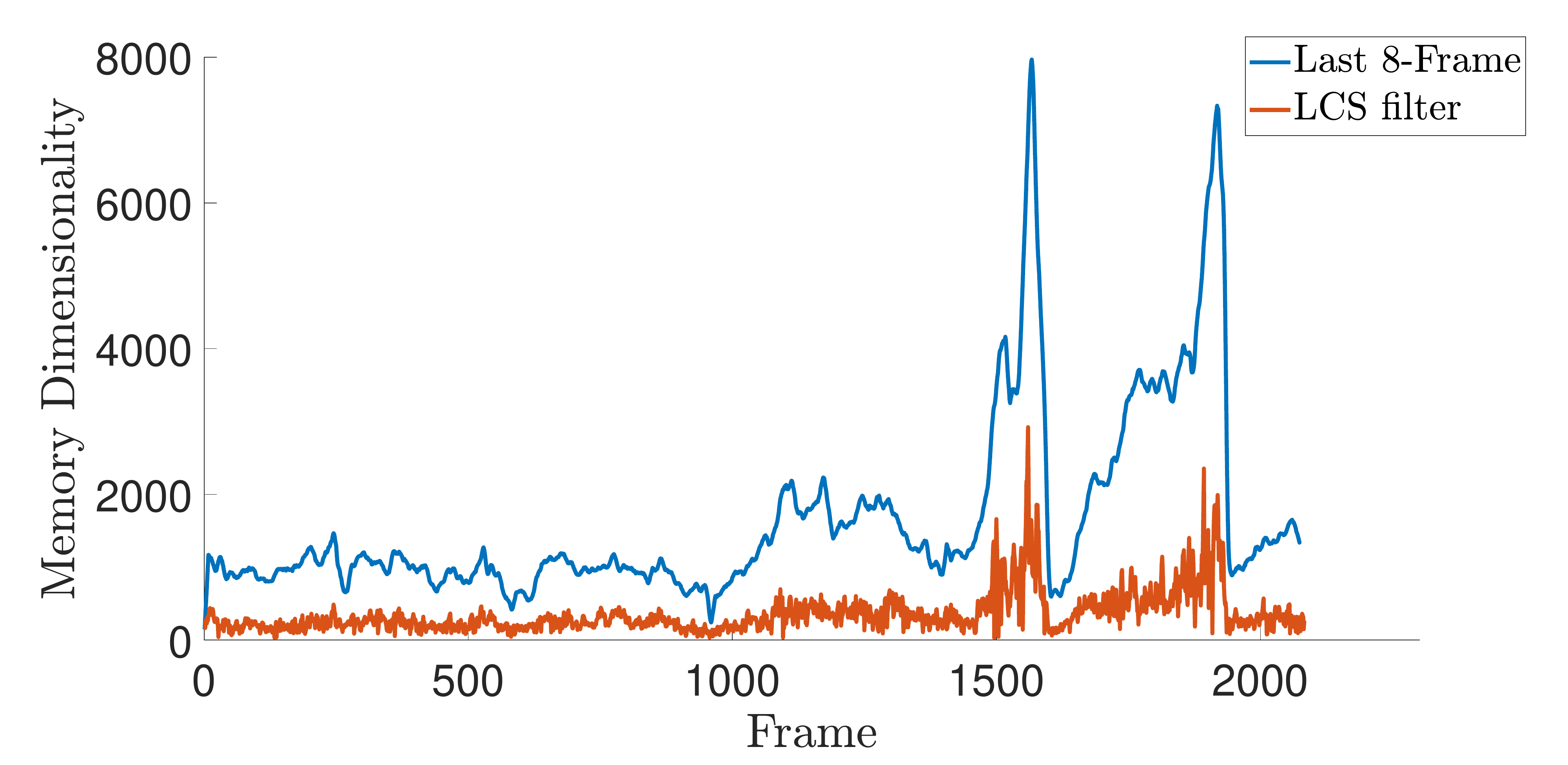}
	\caption{\makehighlight{Dimensionality verses the frame index from the autonomous car experiment.}}
	\label{Fig:RAS_Dim_Auto1}
\end{figure}
\begin{figure}
	\centering
    a)\includegraphics[width=2.8 in]{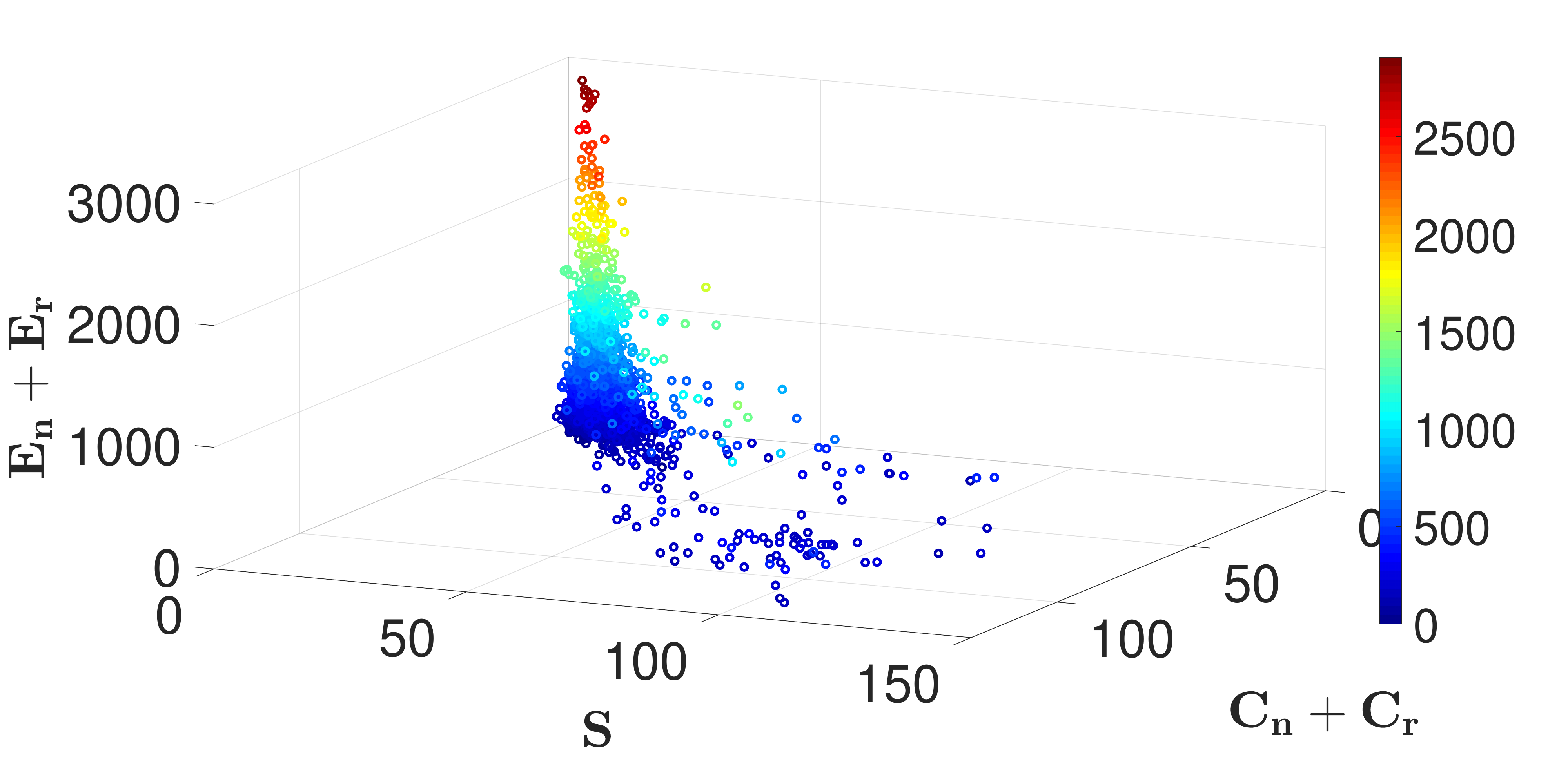}
	b)\includegraphics[width=2.8 in]{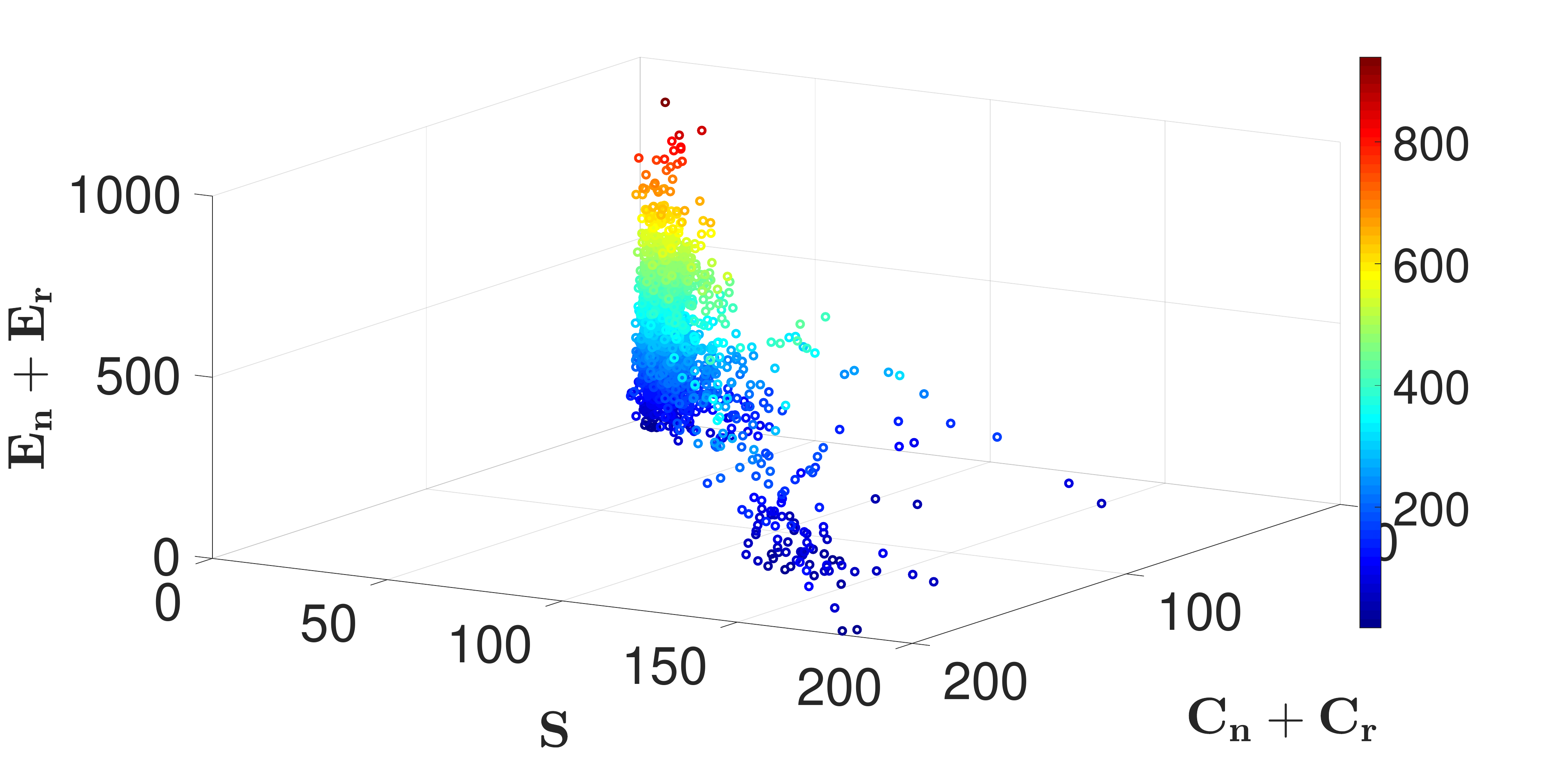}
	\caption{\makehighlight{The dimensionality (memory size) comparison of the filter in each frame. a) It is the scene at day light with no disturbance on camera. b) It is the night scene with raining disturbance at camera lens.}}
	\label{Fig:3dresultRAS}
\end{figure}

\makehighlight{In order to verify our findings in simple scene analysis (dimensionality study) in the previous study (lab environment), the dimensionality of the filter is presented in Figure \ref{Fig:RAS_Dim_Auto1}. The filter outperforms the last 8-frame with an average of 360 overall sizes of which shows how effectively our filter can decrease the memory size even when there is overwhelmingly large numbers of landmarks are detected. Note that we have excluded accumulative memory save since the size of the landmarks rises incredibly large (can be understood from the last 8-frame case). Also, the consistency of detected edges can be understood from the steady value of the dimensionality. Next, Figure \ref{Fig:3dresultRAS} illustrates the behavior of the proposed filter in general. We see that the filter has a consistent behavior despite the change in the number of landmarks in $\uvec{E}_n+\uvec{E}_r$. The pattern kept almost the same in the worst-case scenario as Figure \ref{Fig:3dresultRAS}-b. However, the droplets that disturbed the motion field prediction through the filter have deviations which that is why the number of circles and squares are more. Overall, we think the filter is one of its kind that can operate at an acceptable level in highly disturbed images that were not normally considered in previous studies.   }

%% file: contents/conclusion.tex
\section{Conclusion} \label{sec:conclusion}
In this paper, we present a novel geometric Line-Circle-Sqaure (LCS) filter that is the successor of the original Line-Circle (LC) filter \cite{tafrishi2017line}. 
This filter has three layers, with an additional square layer compared with the LC filter, each of which consists of experts that have dedicated purposes. The first expert is the Line expert to collect, ignore and group the edges depending on the incoming feed from other experts. Next expert, the Circle expert, estimates the edges with defined kinematics, and then it groups the edges as a layer (circles) depending on their locations, angle, and velocity in the frame. Finally, the Square expert collects and matches the circles with geometrical and kinematic conditions. The Square expert with the highest information level tries to detect the objects partially or fully. Also, there is information feed between experts to keep them updated and minimize the level of overconfidence.  With the obtained filter, the data of the landmarks becomes more accurate, compact, and dynamic as more frames are collected from the scene.

The main advantage of the LCS filter is to provide accurate features of the environment at different levels from low (edges) to high (layers). Also, it reduces the computation with its defined learning feed when there are overwhelming features in the environments. It is important to note that this filter is not meant to recognize objects with high accuracy. However, the proposed LCS filter can be utilized as the supervisor to feed only the specific regions of the frame that conventional recognition approaches will determine the objects. On the other hand, this filter without having any pool of images or reference images is able to detect and track objects fully or partially (layers) depending on their kinematics and amount of information that comes from the captured images.



For future work, we will investigate parallel execution of the experts of the filter with utilizing multi-core processors to speed-up the computation. Additionally, we plan to apply the error estimation matrix to all experts for operating the filter faster and more accurately in our incoming works. We think that the proposed filter will be a high potential candidate after the integration of error estimation to the decision parameters. This will help us to automate the choice of the best trust factors in the air within highly dynamic scenes.